\documentclass[10pt,journal,compsoc]{IEEEtran}
\ifCLASSOPTIONcompsoc
  \usepackage[nocompress]{cite}
\else
  \usepackage{cite}
\fi
\usepackage{amsmath}
\DeclareMathOperator*{\argmaxA}{arg\,max} 
\usepackage{multirow}
\usepackage{comment}
\ifCLASSINFOpdf
   \usepackage[pdftex]{graphicx}
   \usepackage{caption}
\usepackage{amsfonts}
\usepackage{bm}
\usepackage{amssymb}
\else
\fi

\begin{document}
%
\title{Robust Pooling Through The Data Mode}
%
%
%
%

\author{\uppercase{Ayman Mukhaimar},
\uppercase{Ruwan Tennakoon, Chow Yin Lai, Reza Hoseinnezhad, and Alireza Bab-Hadiashar},
}
\author{\IEEEauthorblockN{Ayman Mukhaimar\IEEEauthorrefmark{1},
Ruwan Tennakoon\IEEEauthorrefmark{2},
Chow Yin Lai\IEEEauthorrefmark{3}, 
Reza Hoseinnezhad\IEEEauthorrefmark{1} and
Alireza Bab-Hadiashar\IEEEauthorrefmark{1}} \\
    \IEEEauthorblockA{\IEEEauthorrefmark{1}School of Engineering, 
RMIT University, Melbourne, Australia}\\
\IEEEauthorblockA{\IEEEauthorrefmark{2}School of Science,
RMIT University, Melbourne, Australia}\\
\IEEEauthorblockA{\IEEEauthorrefmark{3} Department of Electronic \& Electrical Engineering, University College London, UK
}
\thanks{Manuscript received April 19, 2005; revised August 26, 2015.}}

%
%

\markboth{Journal of \LaTeX\ Class Files,~Vol.~14, No.~8, August~2015}%
{Shell \MakeLowercase{\textit{et al.}}: Bare Demo of IEEEtran.cls for Computer Society Journals}
%



\IEEEtitleabstractindextext{%
\begin{abstract}

The task of learning from point cloud data is always challenging due to the often occurrence of noise and outliers in the data. Such data inaccuracies can significantly influence the performance of state of the art deep learning networks and their ability to classify or segment objects. While there are some robust deep learning approaches, they are computationally too expensive for real-time applications. This paper proposes a deep learning solution that includes a novel robust pooling layer which greatly enhances network robustness and perform significantly faster than state-of-the-art approaches. The proposed pooling layer looks for data a mode/cluster using two methods, RANSAC, and histogram, as clusters are indicative of models. We tested the pooling layer into frameworks such as Point-based and graph-based neural networks, and the tests showed enhanced robustness as compared to robust state-of-the-art methods.

\end{abstract}

}

\maketitle

\IEEEdisplaynontitleabstractindextext

%
\IEEEpeerreviewmaketitle

\IEEEraisesectionheading{\section{Introduction}\label{sec:introduction}}

%
%
%
%
\IEEEPARstart{T}{he} use of deep learning for several 3D task such as point cloud classification \cite{wu20153d,qi2017pointnet,esteves2018learning,su2015multi,klokov2017escape,Ramasinghe}, retrieval \cite{wu20153d,Ramasinghe}, and segmentation \cite{qi2017pointnet++,wang2019dynamic,qi2017pointnet} has shown great success in recent years. However, the success has largely been confined to 3D CAD based benchmarks such as ModelNet~\cite{wu20153d}, McGill~\cite{siddiqi2008retrieving}, and Shapenet~\cite{chang2015shapenet} with very clean data. Working with 3D point clouds of real scenes where data are inaccurate and may be corrupted by outliers remains a challenge, and real 3D training data for natural and man-made objects are still scarce. Testing recent deep networks on 3D CAD models perturbed with outliers and noise showed that the data perturbation has a huge influence on the classification performance \cite{mukhaimar2019comparative}. One approach to resolve this issue is to train the network with outliers. However, noise and outliers, by definition, are not predictable and it would be difficult to train the network for all possible scenarios. Another approach is to build a robust framework that can diminish the influence of outliers compared to conventional deep networks~\cite{gould2019deep}. This approach has received significant interests in recent years \cite{mukhaimar2019pl,gould2019deep}.

The main issue with robust networks for 3D point cloud processing is their computational requirements~\cite{mukhaimar2019pl,gould2019deep}. For instance, the Pl-Net3D~\cite{mukhaimar2019pl} uses the concept of primitive detection, a well-known robust approach, for classification. The method detects planar geometries in objects and uses their information (dimension, location, inliers ratio) for shape classification. The other well known method, called deep declarative network(DDN)~\cite{gould2019deep}, enables the use of an optimization method, such as M-estimator, within the deep learning network. These approaches are computationally expensive and are not practical for most computer vision applications. 

\begin{figure}
\centering
\includegraphics{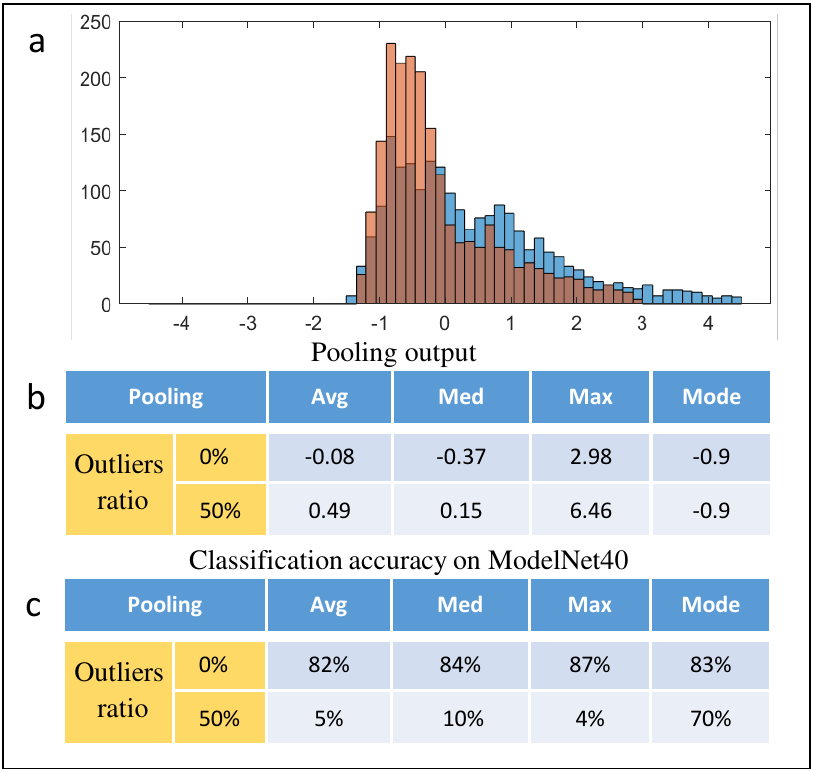}
\caption{ (a) Histogram of the feature vectors (before pooling layer) of the PointNet for clean (orange) and augmented with 50\% outliers (blue) data. (b) The output of the average, median, and max pooing layers. (c) Classification accuracy for the given pooling layers}
\label{fig_1}
\end{figure}

To demonstrate the effect of data perturbation on deep networks performances, we show the effect of the existence of outliers on the PointNet~\cite{qi2017pointnet} by comparing histograms of generated features and the performance of the network as well as its pooling operations on original and augmented (with 50\% outliers) data. The results are shown in figure~\ref{fig_1}. The input of the pooling operation is a feature vector produced by the convolution parts of the network (with a dimension of 1X2048). The figure shows the outputs and performances of the Max, average and Median pooling operations. The results show that the outcomes of max pooling operation for clean and perturbed data are significantly different, a fact which is also reflected in the network's overall classification accuracy (see figure~\ref{fig_1}-c). Not surprisingly, the performance of max pooling, while being ideal for clean data, is most sensitive to the existence of outliers. Interestingly, the figure also shows that using the mean and the median of the data do not diminish the effect of outliers. As such, a better pooling operation is needed to diminish the effect of data inaccuracies on 3D data processing. 


The use of M-estimators as part of the Deep Declarative Network implementation showed promising results for perturbed data~\cite{gould2019deep}. The Truncated quadratic (TQ) or WELSCH (W) estimators-based pooling were able to achieve significantly better robustness to outliers when the PointNet was trained with and without outliers. When an M-estimator is used in a pooling layer, it estimates the location of the mode of feature data. The M-estimator outcomes are shown in figure~\ref{fig_1}-(a), where the mode location is -0.9, for both clean data and augmented (with outliers) data.

Inspired by Deep Declarative Network design, we investigated the use of faster algorithms to perform robust pooling. First, we investigated the use of RANSAC~\cite{fischler1981random} framework for finding the location of the mode. We developed a fast robust method that eliminates the need for the iterative procedure of the conventional RANSAC. In theory, both M-estimators and RANSAC are looking for high cluster regions, and both should give similar results. However, our experiments showed that RANSAC achieved better performance and robustness, which could be related to the existence of isolated non-convex regions of the cost functions during optimization of 'TQ' and 'W' estimators. As conventional RANSAC implementation for finding the regions of maximum densities is computationally expensive and grows exponentially with the number of data, we propose a histogram-based pooling layer for finding data modes/clusters, which is shown to be significantly faster than the the above methods.

The histogram pooling layer divides the feature data into uniform regions and selects regions with maximum densities. In theory, the histogram is similar to both RANSAC and M-estimators, where bin size is somewhat equivalent to the inlier/outlier threshold of RANSAC or tuning parameters of M-estimators. But unlike both approaches, using the histogram mode is significantly cheaper and enables the network to be used for real-time applications. The testing time was found to be around 150 times faster than both RANSAC or an m-estimator. Moreover, histogram`s computational complexity is almost invariant to data size, while it grows exponentially for RANSAC. Importantly, extensive set of experiments on both clean and perturbed data show that the proposed method, while being significantly faster, has higher classification accuracy and outperforms competing methods.

The rest of this paper is structured as follows: We first discuss the latest deep learning classification networks. We then explain the intuition behind using data mode instead of maximum or average pooling as well as the inner working of the proposed RANSAC and histogram-based pooling frameworks. This is followed by the analysis of the performance of the proposed pooling layers under different data augmentation in Section~\ref{sec:Experiments}. Section~\ref{sec:ABLATION STUDY} present a sensitivity analysis of network parameters, followed by Section~\ref{sec:con}, which concludes the paper.


\section{Related work}
\label{sec:Related work}

\subsection{Point cloud classification}

Deep learning frameworks for 3D point cloud classification can be grouped into: Multi-view CNNs (MV-CNN)~\cite{su2015multi,shi2015deeppano,sfikas2017exploiting,kanezaki2018rotationnet,yu2018multi,johns2016pairwise,bai2016gift,sfikas2018ensemble,su2018deeper,kim2020triplanar}, 3D-CNNs~\cite{Wu2015,riegler2017octnet,Vishakh2016a,wang2017cnn,maturana2015voxnet,zhou2018voxelnet,xiang2019novel}, direct~\cite{qi2017pointnet,qi2017pointnet++,Chen_2019_CVPR,Zhang2019RotationIC}, and feature based~\cite{klokov2017escape,li2018so,mukhaimar2019pl} methods. MV-CNN~\cite{su2015multi} usees a preprossessing step for projecting 3D objects from different views into 2D images and feeding those into separate 2D-CNNs. A max pooling layer aggregates features of all the 2D-CNNs and makes the network to be invariant to the view angle. The advantage of this method is that it can use a pretrained CNN module to boost the classification performance. Similarly, there are approaches that use panoramic images of the 3D objects to capture their entire surfaces \cite{shi2015deeppano,sfikas2017exploiting,sfikas2018ensemble}. 

The second group uses 3D-CNNs to classify voxel-based objects. Partitioning the space into voxels based on the point cloud of objects provides a sorted input for a 3D-CNN. Initial approaches were restricted to a grid size of 30 X 30 X 30 due to the large memory and computation power required for 3D-CNNs, However this problem was overcame by the use of octree-grids \cite{wang2017cnn,riegler2017octnet} and spherical use of 3D-CNNs.

The third group directly works on unsorted point cloud as an input to the network layers such as PointNet~\cite{qi2017pointnet}. Most of the methods in this approach use a deep Multi-layer perceptron (MLP) network and a pooling layer as part of their architectures. Such a network is invariant to the order of the data in a 3D point cloud. In related approach, DGCNN~\cite{wang2019dynamic} uses a graph CNN to capture local features in the point cloud. While PointNet applies a pointwise convolution, DGCNN include neighbouring points in its convolution operations. 
Another approach in this group uses a hierarchical network structure as seen in PointNet++~\cite{qi2017pointnet++}, where at each stage of the structure, convolution operations produces new sets of features from input point clouds. The new features are then fed to the next layer of the hierarchical network. In this approach, local regions are determined using farthest point sampling and their features are estimated using PointNet. Unlike PointNet, the proposed structure enables the network to learn the local structures in the point cloud data. DensePoint~\cite{liu2019densepoint} uses PointNet++ framework as well as applying a multi-scale neighborhood segmentation to learn point density information for different size of locality. RS-CNN~\cite{liu2019relation} also uses PointNet++++ framework but local information are learned based on points relative positions as well as the multi-scale neighborhood regions. PVCNN~\cite{liu2019point} combines the PointNet approach with voxel-based neural network models. The combined model was shown to have advantages in terms of memory usage and computational efficiency compared to the original PointNet. \\


The last group employs features detection algorithms such as self-organized-maps of SO-Net~\cite{li2018so} or Kd tree in Kd-Net~\cite{klokov2017escape} to capture the features required for object recognition. Self-organized-maps gather spatial distribution or features in the point cloud while the Kd tree in Kd-Net~\cite{klokov2017escape} captures the shape descriptors to perform deep learning. \\

The robustness of a number of these networks, from the different categories, was examined extensively \cite{mukhaimar2019comparative}. The study compared PointNet, PointNet++, Kd-Net, Oct-Net, and MV-CNN for data with outliers, noise, and missing points. The study showed that the classification performance of MV-CNN and PointNet++ was heavily affected by the above data augmentation forms. As MV-CNN uses several 2D images of a 3D model (i.e. 70 images captured from different locations), outliers would appear in most images, and overall number of outliers is magnified by the imaging process (70 times). PointNet showed good performance against noise and missing points, but was heavily affected by outliers. Both PoinNet and PointNet++ use max pooling as part of their networks, which causes the networks to select outliers as maximum values during their pooling operations. The Kd-Net was heavily affected by outliers as the existence of outliers changes the structure of the Kd-tree graph that performs the classification. Oct-Net showed good level of robustness to noise and outliers, but was affected by missing points. We used the same testing framework and tested other methods that are based on PoinNet and PointNet++ such as PVCNN and RS-CNN and our results showed similar performance to the original PoinNet and PointNet++ methods for the augmented data.


\subsection{Robust classification frameworks}

 Although robust methods fundamentally belong to any of the previous groups, we categorize those separately. For instance, the Pl-Net3D~\cite{mukhaimar2019pl} is a feature based method that combines a primitive fitting technique with PointNet to achieve robust classification. The method employs RANSAC to find instances of geometric primitives in 3D point clouds. The features of those primitives are then used to classify objects. Deep declarative networks~\cite{gould2019deep} proposes a framework for optimization methods to be implemented as part of deep learning networks. The proposed framework allows robust statistical approaches such as M-estimators \cite{leroy1987robust} to be implemented in deep learning. Both Pl-Net3D and M-estimators involve significant computation and high processing time, limiting their ability for real time usage. In this paper, we present a robust classification framework that competes with the state-of-art methods in terms of accuracy, robustness and computational load.

\section{Method}
\label{sec:Method}

\begin{figure}
\centering
\includegraphics[scale=.8]{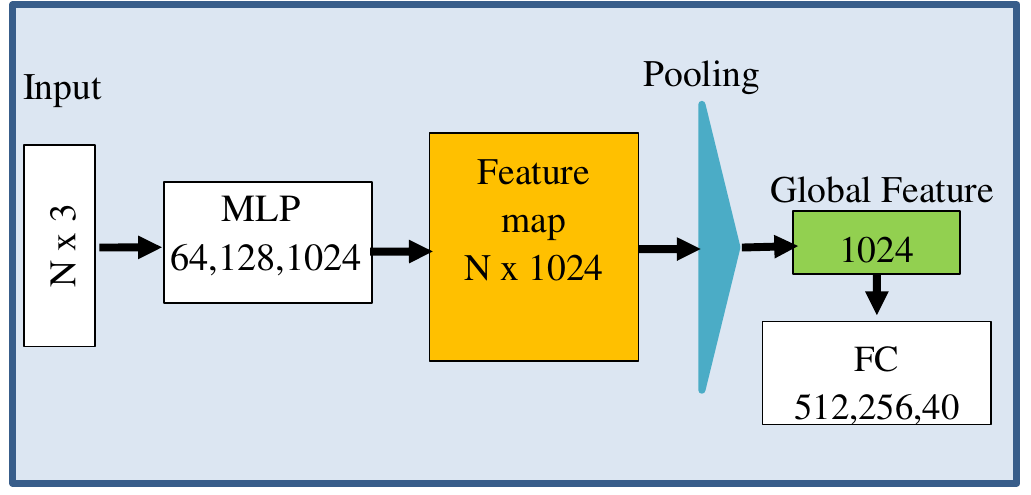}
\caption{PointNet~\cite{qi2017pointnet} Architecture, Input point cloud is forwarded to layers of MLP to generate the n x 1024 feature map in orange, pooling operation produce the global feature in green, followed by Fully connected layers.}
\label{fig_model}
\end{figure}

Given a point cloud of an object that contains outliers, our objective is to build a robust classification deep learning network. To achieve robustness, we propose to use a robust pooling operation in our network. We introduced robust pooling to PointNet~\cite{qi2017pointnet} and DGCNN~\cite{wang2019dynamic} architectures as both methods use global pooling operations. PointNet architecture is shown in figure~\ref{fig_model}. The multi-layer perceptron (MLP) followed by Pooling operation is commonly used in most of point cloud deep learning networks. The architecture represents a symmetric operation on all points, which enable the network to work with unsorted inputs. DGCNN uses similar architecture except that the network accounts for neighboring points. The pooling operation is the main reason for such architecture to fail to classify objects as seen from figure~\ref{fig_1}, hence achieving robust pooling improves the overall robustness.

\subsection{Problem statement}

For the given block diagram shown in Figure~\ref{fig_model}, the feature mapping block (shown in orange colour) provides $N$ feature vectors in the $\mathbb{R}^{1024}$ space. In the dataset used to benchmark the performance of the proposed solution in our paper, typically there are around $N = 2048$ points in the cloud that are mapped (through a MLP) into $N = 1024$ features. 



A major task embedded in the diagram shown in Figure~\ref{fig_model} is the \textit{pooling} task (shown in cyan colour), where it \textit{aggregates} the $N$ vectors and obtain a \textit{single vector} as the \textit{best} representative of the information contents of the those vectors. Let us denote the $i$-th feature vector by 
$$
\bm{x}_i = (x_{i,1}, x_{i,2}, \ldots, x_{i,D})
$$
where $D = 1024$ and $i = 1, 2, \ldots, N$. The $N$ features are then treated as samples of a joint feature density $p: \mathbb{R}^D \rightarrow \mathbb{R}$. The information content of the point cloud is then represented by the entropy of the density that is given by:
\begin{equation}
E[p] \triangleq -\int p(\bm{x})\,d\bm{x}.
\label{eq:entropy_definition}
\end{equation}
Given the samples, we approximate the density as the sum of Dirac delta terms centered at the samples and weighted by their point pdf values:
\begin{equation}
    p(\bm{x}) \approx \sum_{i=1}^N p(\bm{x}_i) \delta(\bm{x} - \bm{x}_i).
\end{equation}
Substituting the above approximation in equation~\eqref{eq:entropy_definition} returns the following approximation for entropy:
\begin{equation}
    E[p] \approx - \sum_{i=1}^N p(\bm{x}_i) \log\left(p(\bm{x}_i)\right).
\end{equation}
Among all the terms included the summation, the largest one is associated with the maximum aposteriori (MAP) estimate of the feature, i.e.
\begin{equation}
    \hat{\bm{x}}_{\mathrm{MAP}} = \arg\max_{\bm{x}} p(\bm{x}).
    \label{eq:MAP_def}
\end{equation}
Therefore, we choose the MAP estimate as the most informative aggregate of all the $N$ feature (the output of pooling operation in Figure~\ref{fig_model}).  

The main problem is how to estimate the MAP. In practice, we are have around $N = 2048$ features in a space dimension of $D = 1024$. In terms of a mesh grid, this is equivalent to having $m = \lceil \log_{D}N \rceil = 2$ bins per dimension in a 1024-D histogram, which is not sufficient for the purpose finding the peak of the joint density. 

Our solution to this problem is to find an approximate for the peak of the very high-dimensional joint density by forming marginal densities ($D$ instances of them, one for each dimension), and locate the peak of each marginal density, separately, then put the coordinates of those peak points together to form the approximate location of the peak of the joint density in the $D$-dimensional feature space. In other words, we find an estimate, named \textit{Marginal MAP} (MMAP) estimate, given by
$$
\hat{\bm{x}}_{\mathrm{MMAP}} = \Big(\hat{x}_{1_{\mathrm{MAP}}},\cdots,\hat{x}_{D_{\mathrm{MAP}}}\Big),
$$
where $\hat{x}_{i_{\mathrm{MAP}}},\ i=1,\ldots,D$, is the MAP estimate for the marginal density in the $i$-th dimension.




In our application, assuming that the MLP and Feature Map networks in Figure~\ref{fig_model} are trained, we expect MMAP and MAP estimates to be reasonably close to each other in such a way that MMAP estimate can be still declared as the aggregate feature that holds substantial amount of information encapsulated in the $N$ feature samples produced by the Feature map block in Figure~\ref{fig_model}.

To explain the intuition behind the above statement, first note that the fully-connected network that inputs the aggregated (global) feature is indeed a mapping from the 1024-D space to 40 different classes of objects. As such, the outputs of the fully-connected network are expected to be very close to one of the coordinate unit vectors $\bm{e}_j = \begin{bmatrix} \bm{0}_{i-1}^\top & 1 & \bm{0}_{40-i}^\top\end{bmatrix}^\top$ where $\bm{0}_{k}$ means a $k$-dimensional vector of zeros. Thus, we intuitively expect that after the network is trained, the global feature input to the fully-connected network ends up in one of 40 different zones in the 1024-D space that are quote distinct. In fact, this is what is expected for the $N$ feature samples; they end up being located together within one of those 40 zones. Hence, with no outlier samples, we expect to see a single-peak distribution of the features (similar to a multivariate Gaussian), and for such a distribution, the peak location (MAP estimate) and the MMAP estimate are very close if not identical. 

\begin{figure}
    \centering
    \begin{tabular}{cc}
        \includegraphics[width=0.22\textwidth]{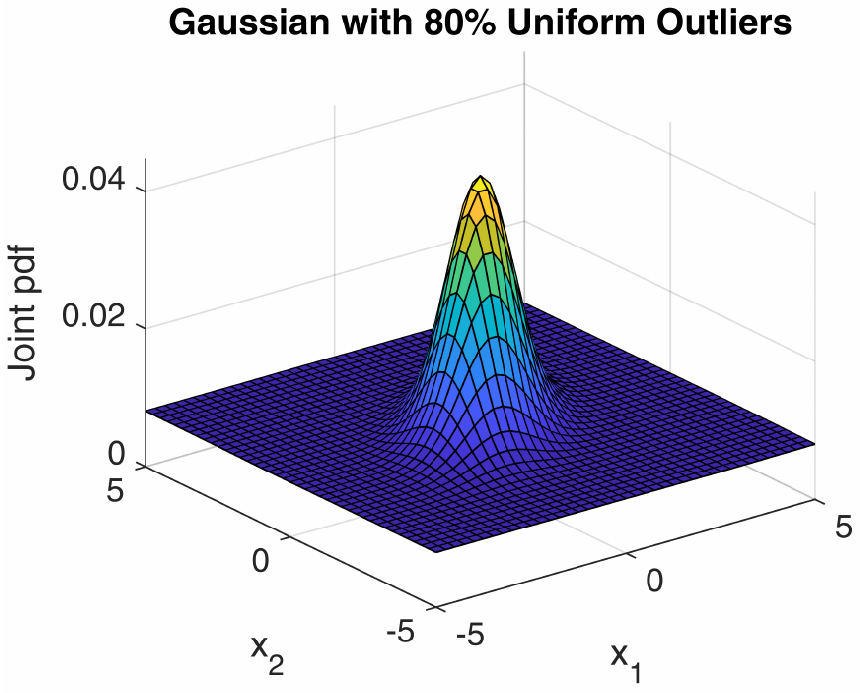} & \includegraphics[width=0.22\textwidth]{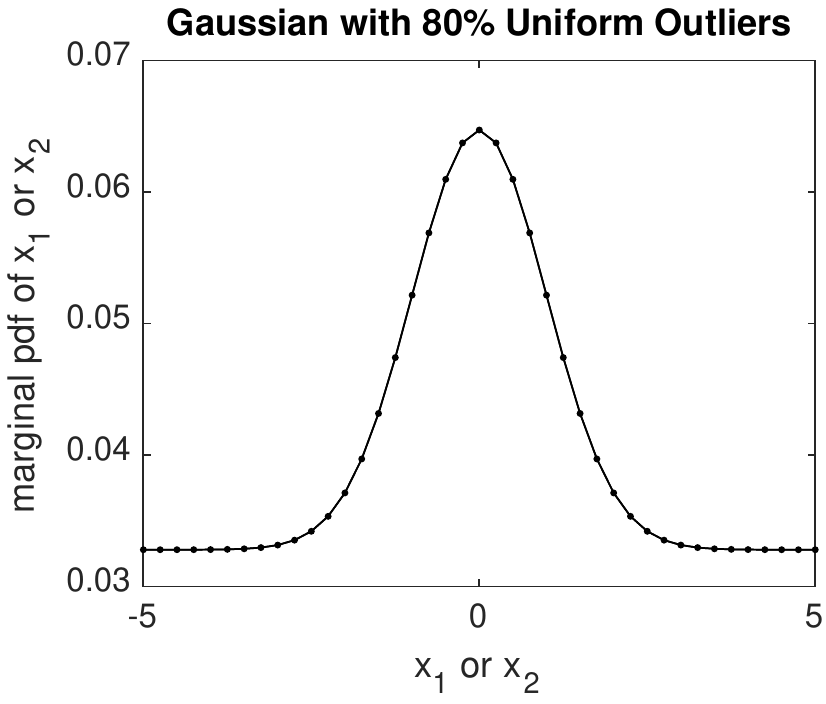} \\
        \small (a) & \small (b)
    \end{tabular}
    \captionsetup{singlelinecheck=off}
    \caption{\textbf{(a)} Density plot of a 2D Gaussian density $\mathcal{N}(\cdot;\bm{\mu}_1,\Sigma_1)$ with $\bm{\mu}_1 = \bm{0}$ and 
    $\Sigma_1 = \big[\begin{smallmatrix} 1 & 0.5 \\ 0.5 & 1 \end{smallmatrix}\big]$ mixed with 80\% outlier samples that are distributed uniformly in $[-5\ 5]\times[-5\ 5]$. \textbf{(b)} Marginal densities of $x_1$ and $x_2$ for the joint density shown in part (a) of this figure.
    }
    \label{fig:2D_densities_1}
\end{figure}

An example is shown in Figure~\ref{fig:2D_densities_1}, demonstrated in 2D for the purpose of visualisation. Figure~\ref{fig:2D_densities_1}(a) presents the density of data comprised of 80\% outlier samples that are uniformly distributed and inliers being distributed according to a joint Gaussian. We observe that the peak is at $[0\ 0]^\top$, while the peaks of the two marginal densities shown in Figure~\ref{fig:2D_densities_1}(b) are both located at zero.

\begin{figure}
    \centering
    \begin{tabular}{cc}
        \includegraphics[width=0.22\textwidth]{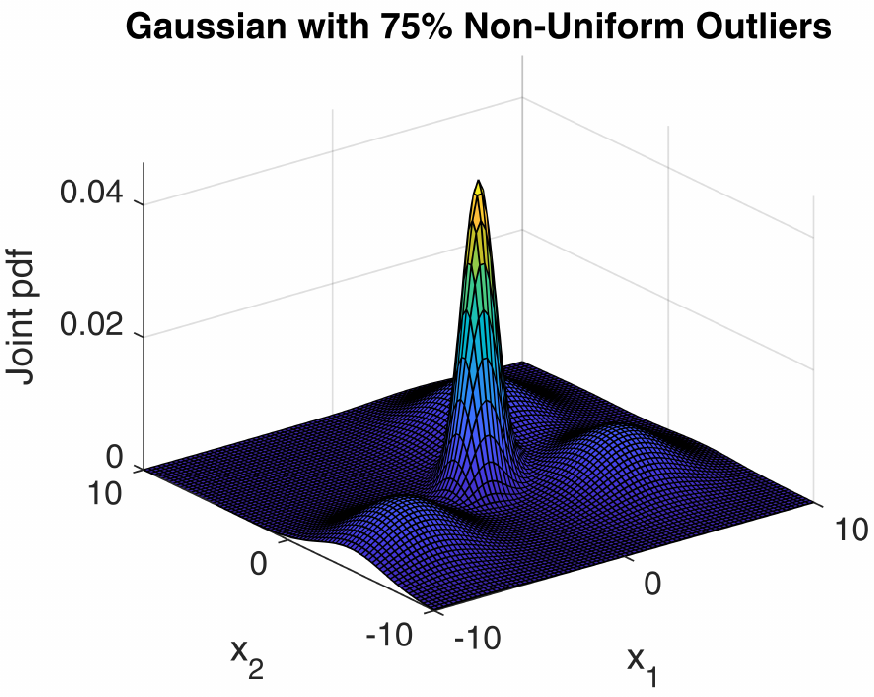} & \includegraphics[width=0.22\textwidth]{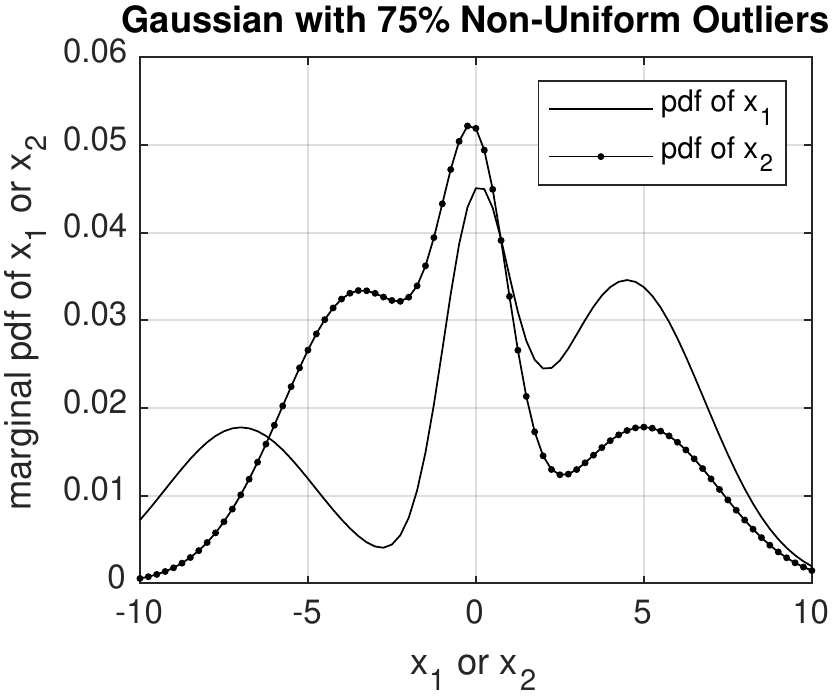} \\
        \small (a) & \small (b)
    \end{tabular}
    \captionsetup{singlelinecheck=off}
    \caption{\textbf{(a)} A Gaussian mixture density in 2D, comprised of four components $\mathcal{N}(\cdot;\bm{\mu}_i,\Sigma_i), i=1,\ldots,4$ with equal weight 0.25. The first component is same as shown in Figure~\ref{fig:2D_densities_1}(a). Parameters of the other components are: 
    $\bm{\mu}_2 = \big[\begin{smallmatrix} 5 \\ 4 \end{smallmatrix}\big]$,
    $\bm{\mu}_3 = \big[\begin{smallmatrix} -3 \\ 5 \end{smallmatrix}\big]$,
    $\bm{\mu}_3 = \big[\begin{smallmatrix} -4 \\ 7 \end{smallmatrix}\big]$,
    $\Sigma_2 = \Sigma_3 = \Sigma_4 = \big[\begin{smallmatrix} 5 & 0.5 \\ 0.5 & 5 \end{smallmatrix}\big].$
    \textbf{(b)} Marginal densities of $x_1$ and $x_2$ for the joint density shown in part (a) of this figure.}
    \label{fig:2D_densities_2}
\end{figure}

If the outliers are not uniformly distributed, as long as they do not themselves form a sharper peak in the density, we still expect the MMAP estimate to be close to the peak location. This is visualised in an example shown in Figure~\ref{fig:2D_densities_2} where 25\% of data are the inliers distributed in a similar way to the previous case, and the rest of data (outliers) are equally scattered around four points. Figure~\ref{fig:2D_densities_2}(b) demonstrates that due to the outliers, the marginal density peaks are slightly deviated from the peak of the combined density.

\subsection{Histogram and RANSAC Pooling}

 Robust fitting techniques aim to find data clusters that represent instances of a given model. To apply RANSAC, a collection of $m$ hypotheses were examined to find the inliers within a threshold $\epsilon$ for all these $m$ hypotheses. The hypothesis with the maximum number of inliers is then chosen as the best model estimate. This translates to looking for:


\begin{equation}
\hat{m} = \argmaxA_m  ( \sum\limits_{i=0}^{N}  |x_{i}-x_{m}| \leq \epsilon )
  \label{equ:ran}
\end{equation}
where $\hat{m}$ corresponds to the point with maximum number of inliers (equivalent to $\hat{\bm{x}}_{\mathrm{MAP}}$) and the output of the pooling layer in the forward step.

For one dimensional data, a histogram can be viewed as a density estimator where data is partitioned into intervals (bins) and their density is estimated by counting the number of data in a bin. We use a histogram as part of our proposed pooling operation, in which $\hat{L}$ is the index of the mode bin $\hat{L} = \argmaxA_m  p(L_{i})$, where $L \subset R^m$ is a set of bin indices for $m$ bins. In comparison to RANSAC for location estimation, the bin size is equivalent to the threshold $\epsilon$ and the histogram mode is the equivalent to the model with maximum number of inliers. 



\subsection{Robustness of mode pooling }

\begin{figure}[h]
\centering
\includegraphics[scale=.25]{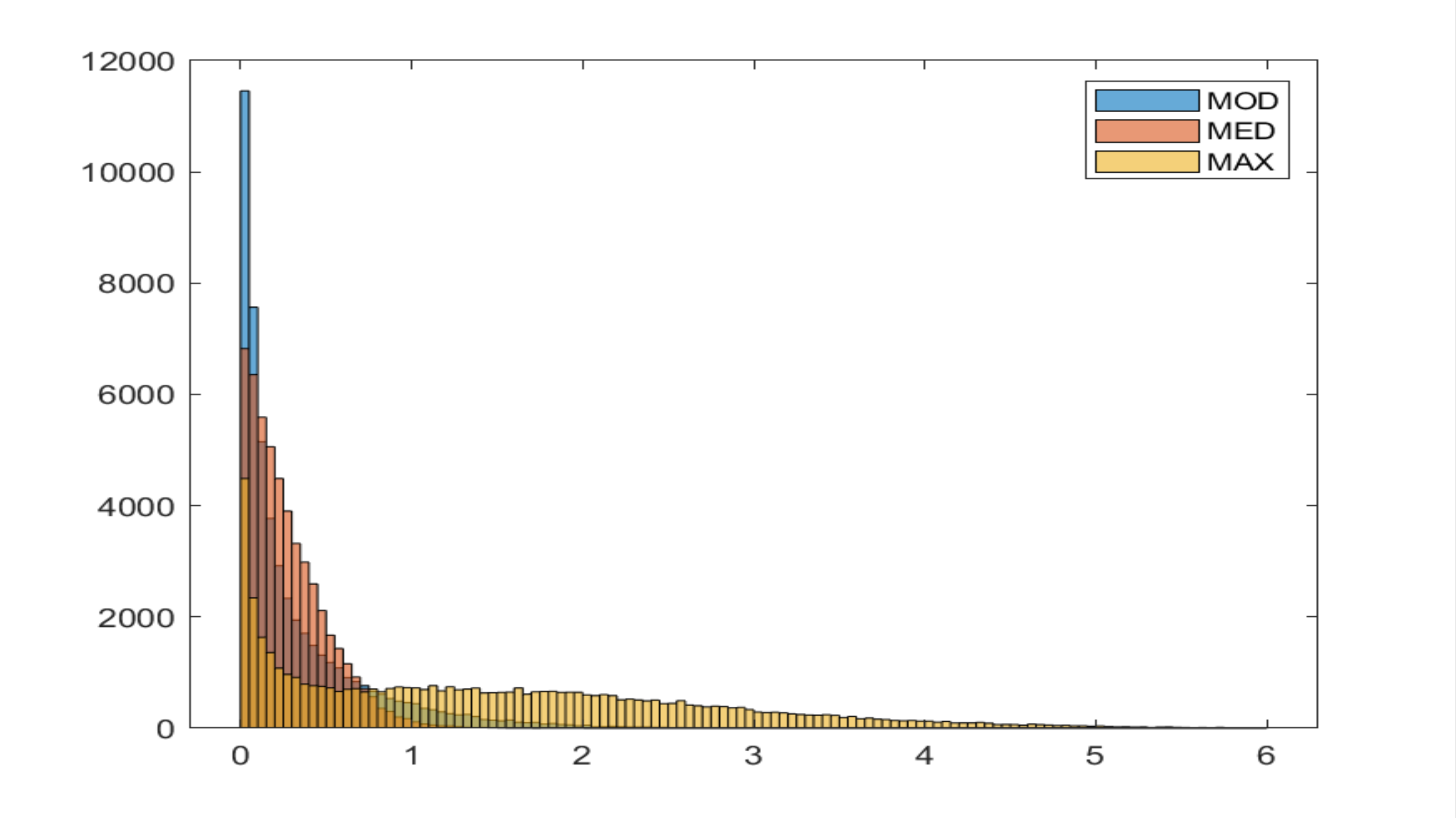}
\caption{Difference in pooling output between features of the clean and its outlier augmented point clouds of an object for mode, max, and median pooling operations.}
\label{fig_his}
\end{figure}

To compare the robustness of the max, mean, median, and mode in pooling operations, we first randomly selected $50$ (out of $2048$) feature vectors (of size $1024$) of the feature map (shown in figure~\ref{fig_model} by the orange box). The experiment was repeated for both a clean and an object augmented with 50\% outliers. The above mentioned pooling operations were applied to both feature collections. Pooling outputs of the clean object were subtracted from outputs of the augmented object and the differences were plotted as shown in figure~\ref{fig_his}. Average and median pooling were very similar and the median is only plotted. The figure shows that mode pooling has the lowest output difference between clean and augmented data, indicating significant robustness to presence of outliers. 

\section{Experiments} \label{sec:Experiments}

In this section, we present a comparative analysis of the performance of the proposed pooling operations for different types of data augmentations. We outline the composition of the datasets as well as the network architectures.

\subsection{Datasets}

\begin{figure}

    \centering

 \begin{tabular}{c c c}
 
  \includegraphics[width=0.1\textwidth]{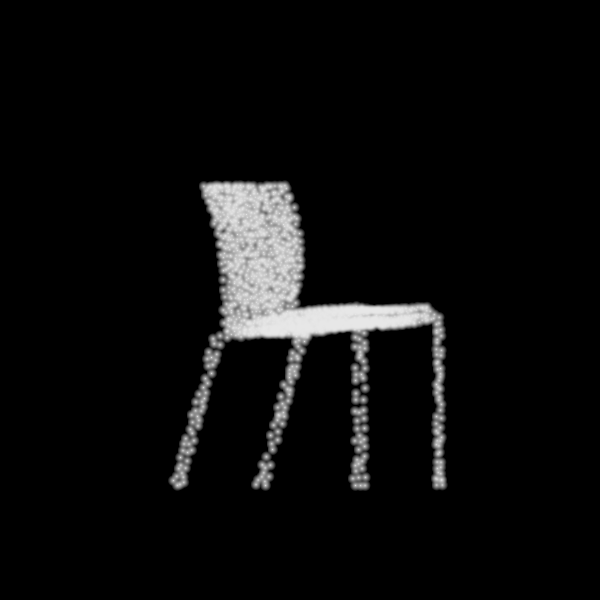} &
    \includegraphics[width=0.1\textwidth]{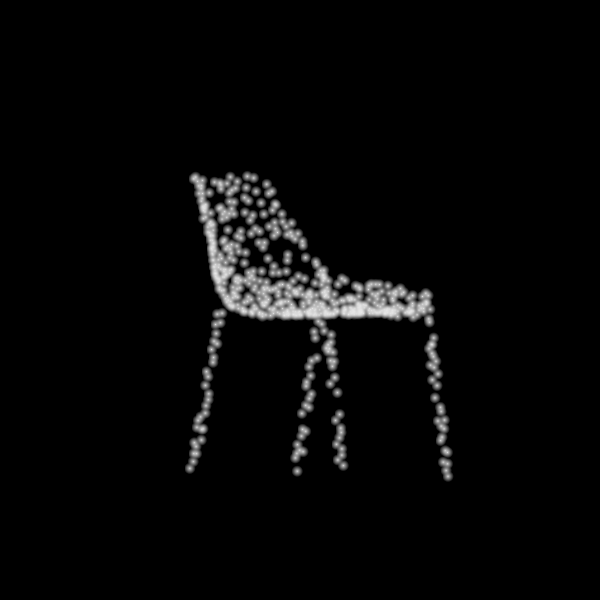} &
    \includegraphics[width=0.1\textwidth]{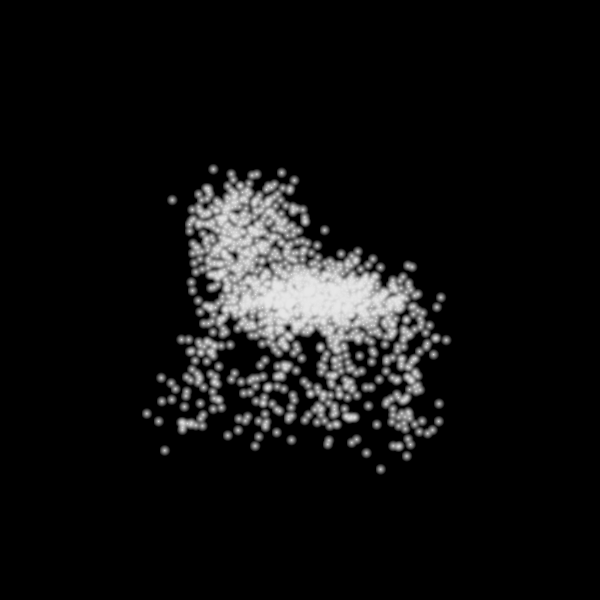}  \\
        \small (a) & \small (b) & \small (c)
   
  \\
  
  \includegraphics[width=0.1\textwidth]{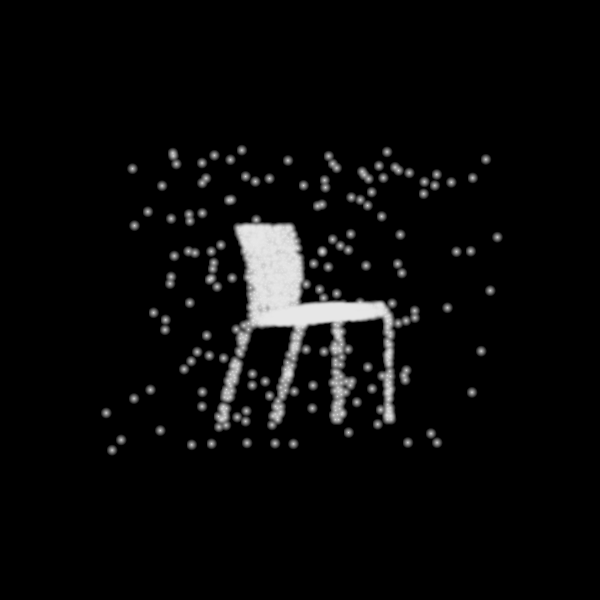} &
      \includegraphics[width=0.08\textwidth]{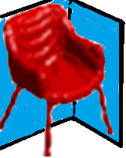} &
  
  \includegraphics[width=0.07\textwidth]{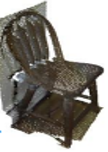} 
\\
        \small (d) & \small (e) & \small (f)

\end{tabular}
\caption{\textbf{(a)} The point cloud of a chair taken from the ModelNet40 dataset, \textbf{(b)} the same chair is augmented with random point dropout,  \textbf{(c)} the same chair is augmented with Gaussian noise, \textbf{(d)} the same chair is augmented with scattered outliers, and \textbf{(e)} the same chair is augmented with clustered outliers which appear as the surfaces in blue. \textbf{(f)} The point cloud of a chair taken from ScanObjectNN dataset including background data (used also as clustered outliers).}
\label{fig_ag}
   
\end{figure}

For classification, we use the ModelNet40 dataset ~\cite{wu20153d,qi2017pointnet} which consists of 9,843 training and 2468 testing samples from 40 categories. Each sample consists of 2048 points normalized within the unit cube. We don't introduce any augmentation to the training data except random rotations, but data augmentations such as noise, random point dropout, and outliers, are introduced to the testing samples. Examples of those augmentations are seen in figure~\ref{fig_ag}. If a point normal is used, we calculate the normal by using twenty of its neighboring points. We also use the ScanObjectNN dataset~\cite{uy2019revisiting} to test the performance of our proposed pooling operations on real-scene data. The ScanObjectNN dataset contains 2902 scenes of objects categorized into 15 categories. Each scene carries the point cloud of an object in addition to the point cloud of background elements or parts of nearby objects as seen in  figure~\ref{fig_ag}. During training, we use the point cloud of objects only, while when testing, we include the point cloud of background and parts of nearby objects as real-scenarios outliers. 

For part segmentation, we use ShapeNet part dataset \cite{yi2016scalable}, which consists of 16,881 shapes from 16 categories, with 50 parts in total. All images are annotated with their parts labels. To examine the robustness, we augmented the test set with random outliers, and when testing, we only used inlier points to calculate the average mIoU.

\subsection{Selected architectures}
To analyze the performance of the proposed pooling operation, we used the PointNet architecture as its multi-layer perceptrons and global pooling are shared by many recent deep learning frameworks. For classification with PointNet, three layers of MLP were used with 64,128,1024 filters respectively. The number of bins for histogram pooling was set to 70 and their centers were uniformly distributed between -10 to 10. RANSAC was implemented with an equivalent threshold of 0.143 and with number of hypothesis $m$ ranges between 30\% to 50\% of the total number of points. The learning rate was set to 0.0001 and number of epochs was set to 100. For ScanObjectNN, two layers of MLP were used with 128 and 4048 filters respectively, and the number of bins for histogram pooling was set to 200.

To examine the effect of the proposed pooling on graph neural networks, we also investigate DGCNN in the sensitivity analysis section (see section~\label{sec:ABLATION STUDY} for detail).

For part segmentation, we used the original PointNet segmentation architecture with the proposed histogram pooling. The number of bins in the histogram was set to 1200 for the interval between -5 to 5. The initial learning rate was set to 0.0001, and the number of epochs was set to 100. 


\subsection{Classification Performance on clean data}

We calculated the classification accuracy for PointNet model with different pooling operations including max, RANSAC, histogram, and Truncated Quadratic (TQ) on ModelNet40 dataset. The results of these experiments as well as the classification accuracy for sate-of-the-art robust methods such as OctNet~\cite{riegler2017octnet} and PL-Net3D~\cite{mukhaimar2019pl} are shown in table~\ref{Tab:t4}. The classification accuracy when using ``PointNet (vanilla)" with max pooling reaches 87\%, which is 3\% higher than using Truncated Quadratic (TQ) and histogram (HS) poolings. Both TQ and HS pooling operations produce similar classification accuracy to OctNet and PL-Net3D (when using point normals). \\
Table~\ref{Tab:tsc4} shows the classification accuracy when training was done on ModelNet40 and when testing was done on ScanObjectNN. The first column (OBJ) shows the classification accuracy for objects without any augmentation. The T25 suffix denotes a 25\% translation in the bounding box of the captured object. R and S denotes for random rotation and scaling respectively. The first set of results were taken from \cite{uy2019revisiting}, while the last four rows show the classification accuracy of PointNet vanilla with the different pooling operations. The results show that mode pooling (TQ, RN, and HS) has higher classification accuracy than max pooling in the first two columns (except for RN pooling in the T25 case). However, when testing with scaling or rotation, max pooling shows higher performance. The results also show that when training on CAD models and testing on real-world data, mode pooling generalizes better than the other compared networks ( for no scaling or rotation).

\begin{table}[]
\begin{center}

\caption{The classification accuracy on ModelNet40}
\label{Tab:t4}
 \fontsize{10}{12}\selectfont
\begin{tabular}{c|c|c}
\hline
Method                                             & Input                                                                      & MN40 \\
\hline \hline
OctNet   & Voxel                                                                      & 86.5 \\
\hline
PL-Net3D                                           & \multirow{6}{*}{Points}                                                    & 86.6 \\
PointNet  &                                                                            & 89.2 \\
DGCNN &                                                                            & 92.2 \\

PointNet++ &                                                                            & 91.8 \\
PointNet* + Max                                   &                                                                            &   87   \\
PointNet* + TQ                                   &                                                                            &   84   \\
PointNet*+HS (ours)                               &                                                                            & 84   \\ 
\hline
PointNet* + Max                                   &       \multirow{3}{*}{\begin{tabular}[c]{@{}c@{}}Points \\+\\ normals\end{tabular}}                                                                      &   89   \\
PointNet*+TQ                                      &   &      87 \\
PointNet*+HS (ours)                               &                                                                            & 86  \\
\hline

\end{tabular}
   
      \small
      \item * PointNet vanilla, + indicates the used pooling operation.
  
\end{center}
\end{table}

\begin{table}[]
\begin{center}
\caption{The classification accuracy when training on ModelNet40 and testing on ScanObjectNN}
\label{Tab:tsc4}
 \fontsize{10}{12}\selectfont

\begin{tabular}{c|c|c|c|c|c}

\hline
Method                        & OBJ  & T25  & T25R & T50R & T50RS \\ \hline
3DmFV   & 30.9 & 28.4 & 27.2 & 24.5   & 24.9    \\
PointNet & 42.3 & 37.6 & 35.3 & 32.1   & 31.1    \\
SpiderCNN                     & 44.2 & 37.7 & 34.5 & 31.7   & 30.9    \\
PointNet++                    & 43.6 & 37.8 & 37.2 & 33.3   & 32.0    \\
DGCNN                         & 49.6 & 42.4 & \textbf{40.3} & \textbf{36.6}   & \textbf{36.8}    \\
PointCNN                      & 32.2 & 28.7 & 28.1 & 26.4   & 24.6    \\ \hline
HS (ours)                            & \textbf{50.2} & 43.2 & 38.6 & 33     & 33      \\
MAX                           & 47.1 & 43   & 40   & \textbf{36.6 }  & 35.3    \\
RN (ours)                             & 48.6 &   40.4   &   36.4   &    31.7   &   32      \\
TQ                            &  \textbf{50.5}    &  \textbf{43.8}    &   39.6   &  34      &  32.3       \\ \hline

\end{tabular}
\end{center}
\end{table}

Table~\ref{Tab:t5} shows the testing and training times, and the used GPU memory for the PointNet with different pooling operations including max, RANSAC, histogram, and Truncated Quadratic (TQ). For comparison, the feature map (shown by an orange block in figure~\ref{fig_model}) dimensions for all pooling operations were set to '10 x 1024 x 2048' and '10 x 512 x 512' for the sizes of batch, number of points, and number of features, respectively. For RANSAC, the number of hypotheses $m$ was half the number of points. With such number of hypotheses, and for a '10 x512 x 512' tensor, 11Gb of GPU memory was used to train the network. The training time for one epoch was 3 minutes, while its testing time was only 14 seconds. TQ requires only 0.7Gb of GPU memory and double the training time. However, the testing time was 4 times longer than RANSAC. For a tensor with the size '10 x 1024 x 2048', looping was required to use RANSAC on a 12GB GPU, which affected both the training and testing times. Histogram only required 9 seconds for training one epoch and 3 seconds to finish testing, almost 100 times faster than TQ. The testing and training speeds are as fast as using max pooling. These results show that the histogram pooling is significantly faster than the other approaches.

  

\begin{table}[]
\begin{center}
\caption{Pooling operations versus GPU usage, testing and training times for two tensor sizes.}
\label{Tab:t5}
 \fontsize{10}{12}\selectfont 
\begin{tabular}{|c|c|c|c|c|}

\hline
Pooling       & \begin{tabular}[c]{@{}c@{}}GPU\\  usage\end{tabular} & \begin{tabular}[c]{@{}c@{}}Tensor\\  size\end{tabular}              & \begin{tabular}[c]{@{}c@{}}Testing\\  time\end{tabular} & \begin{tabular}[c]{@{}c@{}}Train time\\ (one epoch)\end{tabular} \\ \hline \hline
RN           & 11Gb                                                & 10x
 & 14 s                                  & 3 m                                          \\
 TQ          & 0.7 Gb                                                & 512x512
 & 1 m                                  & 7 m                                          \\ 
 
 \hline
RN        & 11Gb                                               &                   & 10   m                                                  & 60   m                                                           \\ 
TQ  & 1.5Gb                                                &   10x                 & 2   m                                                  & 15   m                                                           \\ 
HS            & 2Gb                                                &    1024x2048                & \textbf{3 s}                                                     & \textbf{9 s }                                                             \\ 
Max           & 2Gb                                                &  & 3 s                                  & 11 s                                          \\ \hline

\end{tabular}

      \small
      \item RN: RANSAC, TQ: truncated quadratic, HS: histogram.

\end{center}
\end{table}



\subsection{Classification robustness to outliers}
\begin{figure}[]
\centering
\includegraphics[scale=.33]{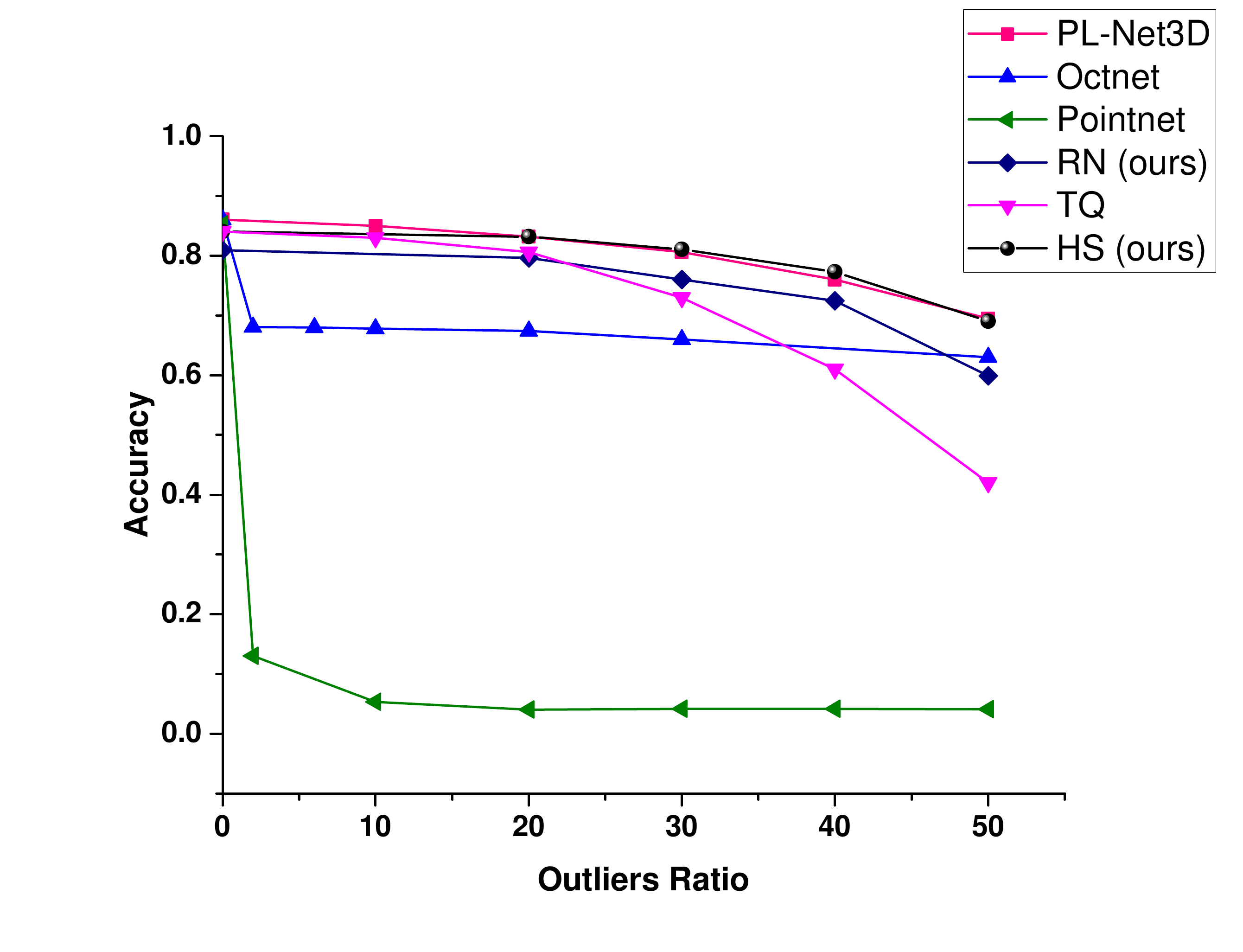}
\caption{Classification accuracy versus outlier ratio.}
\label{fig_outl}
\end{figure}

Visual data often contain outliers as there are imperfections in the scanning methods or processing pipelines such as multi-view reconstruction of 3D models. An example of those outliers is the background elements in the ScanObjectNN dataset. We examined the effect of outliers on the classification accuracy of different techniques and in particular their remaining influence after applying different pooling operations.\\ 
We considered two outliers scenarios, uniformly distributed outliers, and structured outliers (pseudo). In the first scenario, outliers were simulated by adding uniformly distributed points in the unit cube to the ModelNet40 test dataset, with ratios vary from 0 to 50\% of the total number of object's points. We present the results of our experiments with added outliers in figure~\ref{fig_outl}. The tested models are: Oct-net, Pl-Net3D, and PoinNet (vanilla) with several pooling operations including histogram, RANSAC, and Truncated quadratic.

As seen from figure~\ref{fig_outl}, TQ pooling scored a classification accuracy of 40\% at 50\% outliers ratio, while histogram pooling achieved significantly better results with 70\% classification accuracy at the same outliers ratio. The performance of histogram pooling in terms of robustness to outliers is similar to Pl-Net3D, with an advantage of being much faster. The inference time for Pl-Net3D is \textbf{2.7s}, while it is \textbf{0.001s} for our method (about 2000 times faster). The classification accuracy of using Oct-Net and RANSAC pooling at 50\% outliers drops to around 60\%. 

In the second outliers scenario, we test the robustness of our proposed pooling operations on pseudo (structure) outliers. For this expermint, we use both ScanObjectNN and ModelNet40 datasets. The ScanObjectNN dataset contains scenes of objects, where each scene carries the point cloud of an object in addition to the point cloud of background elements or parts of nearby objects. We use the point cloud of background and parts of nearby objects as outliers. We use 1024 points from the original objects, while we use 200, 300, and 400 outlier points (from background elements or parts of nearby objects) as seen in table~\ref{Tab:tt5}. We also use the entire outlier points reported in the dataset. The total number of outlier points reported in the dataset reaches more than 80\% of the original object points in some scenarios. In addition, we generate pseudo (structure) outliers for ModelNet40 dataset by sampling 1024 point from the original objects, and generating two background surfaces near the objects, as seen in figure~\ref{fig_ag}-(e). These generated background surfaces have 200, 300, and 400 outlier as reported for the ScanObjectNN dataset. 

As seen from table~\ref{Tab:tt5}, when training and testing was performed on the clean point cloud of the ScanObjectNN dataset, the classification accuracy of PoinNet vanilla using max, HS, TQ and RN pooling operations reaches 84\%, 80\%, 76\%, and 75\% respectively. While when training on the clean point cloud of the ScanObjectNN dataset and testing on objects with 200 outlier points, the classification accuracy using the max pooling reduces to 75\% (drops by 9\%). However when when using the HS pooling, the classification accuracy only drops by 1\%. The classification accuracy when using TQ and RN pooling operations drop to 73\%. Comparing how much the classification accuracy for each pooling operation drops indicates that TQ, RN, and Hs methods have higher robustness than max pooling. 
When using the entire outliers reported in the dataset, all methods drop by almost 10\%. Figures~\ref{fig_conhs} and ~\ref{fig:conmmx} show the confusion matrices for PointNet with the HS and max pooling operations. Comparing those figures show that the overall classification accuracy when using all outliers mainly drop because of the low classification accuracy of two objects, chairs and tables. The two objects have large number of outliers ratio (ranges between 50\% to 70\%) which could be the reason for the miss-classification, another reason is the high similarity between some objects when outliers exist (i.e. table and desk).   

When using the ModelNet40 dataset that is augmented with pseudo outliers, the classification accuracy of PointNet using max pooling drops from 84\% to around 60\% (drops by almost 20\%) at the different outliers levels shown in table~\ref{Tab:tt5}. While PointNet using TQ only drops by 3\%, 8\%, and 12\% when object have 200, 300, and 400 outliers respectively. RN and HS scores around 2 to 5\% lower classification accuracy than TQ for the different outliers levels. Though TQ scores higher robustness on the augmented ModleNet40 datset, both HS and RN showed higher robustness on the ScanObjectNN dataset (HS scores almost 10\% higher in some cases).






\begin{table*}[]
\begin{center}
\caption{Classification accuracy on ScanObjectNN (OBJ) and ModelNet40 (MN40) for objects with clustered outliers (BG). +100BG means object contains 100 outlier points.}
\label{Tab:tt5}
 \fontsize{10}{12}\selectfont

\begin{tabular}{c|c|c|c|c|c||c|c|c|c|c}
\hline
Method & OBJ & \begin{tabular}[c]{@{}c@{}}OBJ\\+200BG\end{tabular} & \begin{tabular}[c]{@{}c@{}}OBJ\\+300BG\end{tabular} & \begin{tabular}[c]{@{}c@{}}OBJ\\+400BG\end{tabular} & \begin{tabular}[c]{@{}c@{}}OBJ\\+BG\end{tabular} & MN40 & \begin{tabular}[c]{@{}c@{}}MN40\\+200BG\end{tabular} & \begin{tabular}[c]{@{}c@{}}MN40\\+300BG\end{tabular} & \begin{tabular}[c]{@{}c@{}}MN40\\+400BG\end{tabular} \\ \hline
HS     & 80 & \textbf{79 }                                                  & \textbf{79}                                                   & \textbf{77 }                                                  & 70  & 83 & 78 &  73 &  66                                             \\

MAX    & \textbf{84} & 75                                                   & 75                                                   & 74                                                   & \textbf{73}    & \textbf{87} &    61  & 59   &  57                                         \\
RN     & 75 & 73                                                   & 72                                        & 70                                                   & 63     & 81 &  78 & 73 & 66                                                      \\
TQ     & 76 & 73                                                   & 69                                                   & 66                                                   & 58 &  83 & \textbf{80} & \textbf{75} & \textbf{71 }                                           \\ \hline
\end{tabular}

\end{center}

\end{table*}




\subsection{Classification robustness to noise}

\begin{figure}[h]
\centering
\includegraphics[scale=.33]{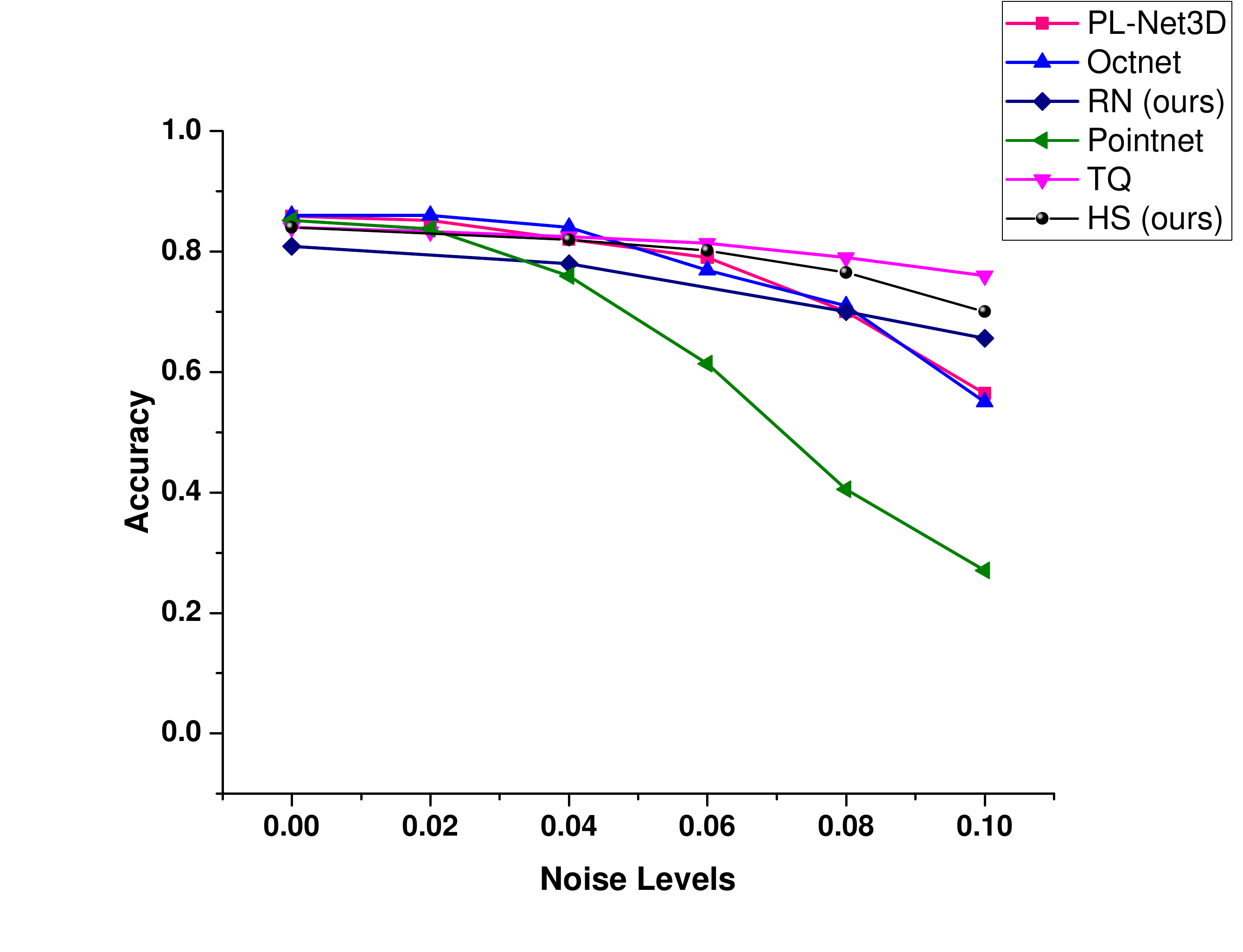}
\caption{Classification accuracy versus noise.}
\label{fig_noise}
\end{figure}

To examine the performance of different pooling operations on the classification accuracy of noisy data, zero mean random perturbations having Gaussian distributions with standard deviations ranging from 2\% to 10\% were added to the testing data. The results of those simulations are shown in Figure~\ref{fig_noise}. TQ and histogram pooling methods outperformed all the other methods. TQ scored classification accuracies of 81\% and 76\% at 0.06 and 0.1 noise levels respectively, while histogram pooling scored classification accuracies of 80\% and 70\% at 0.06 and 0.1 noise levels respectively. Pl-Net3D scored 62\% classification accuracy at 0.1 noise level, followed by Oct-Net with 59\%. RANSAC performance against noise also outperforms Pl-Net3D and Oct-Net with a classification accuracy of 66\% at 0.1 noise level.


\subsection{Classification robustness to random point dropout}

\begin{figure}[h]
\centering
\includegraphics[scale=.33]{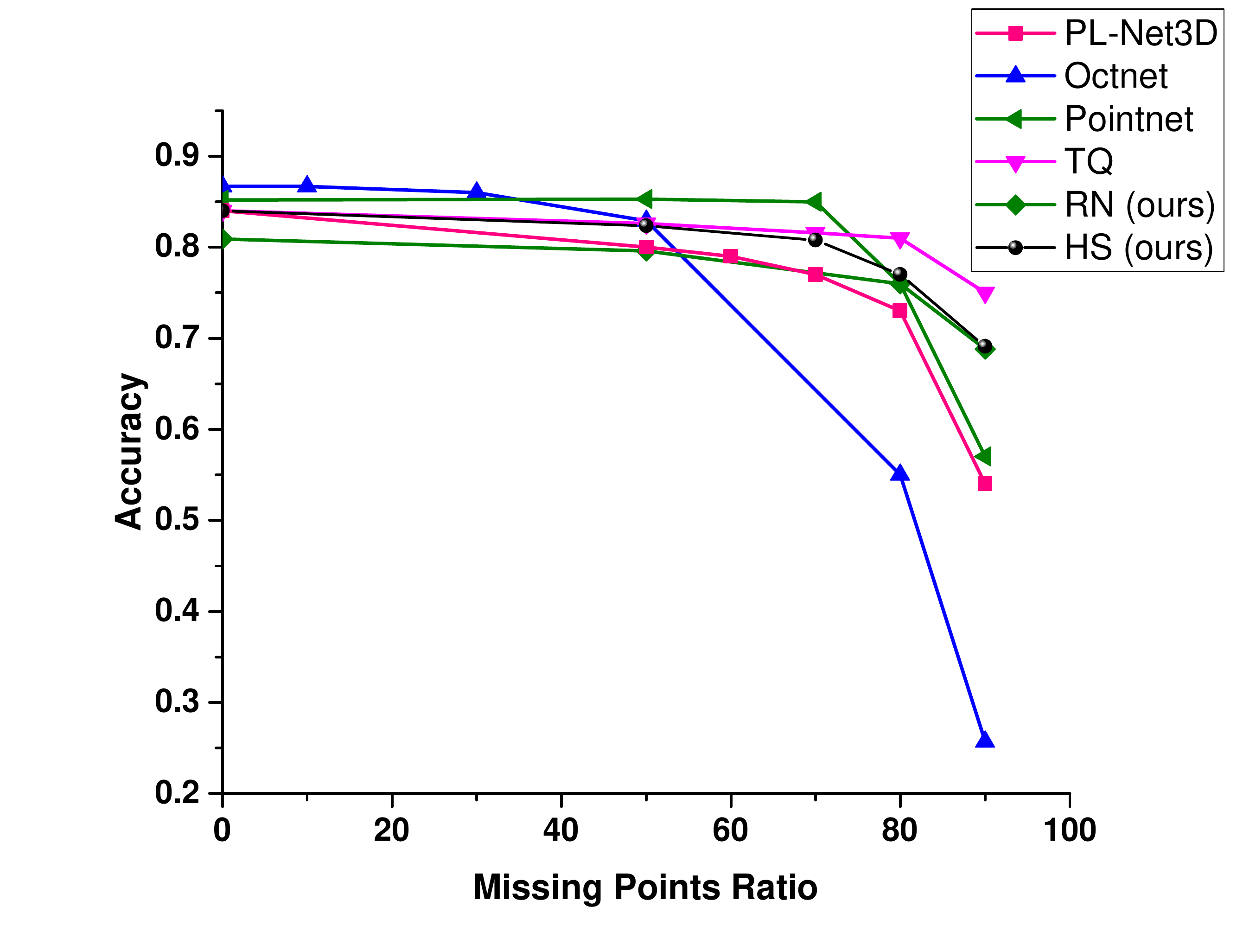}
\caption{Classification accuracy versus missing points.}
\label{fig_missp}
\end{figure}

We performed random point dropout to the testing set with values ranging from 50\% to 90\%. The classification performance of different methods are compared for those augmentation and the results are shown in figure~\ref{fig_missp}. PointNet with max pooling showed the highest robustness up to 70\% random point dropout, however using the TQ pooling showed the highest robustness at the higher percentages, followed by RANSAC and histogram pooling. Both TQ and histogram methods only drop by 1~1.5\% at 50\% points dropout, while Pl-Net3D drops by 2.5\%. OctNet performance deteriorates rapidly after 50\% dropout.

\subsection{Segmentation robustness to outliers}

\begin{figure*}[h]
\centering
\includegraphics[scale=.5]{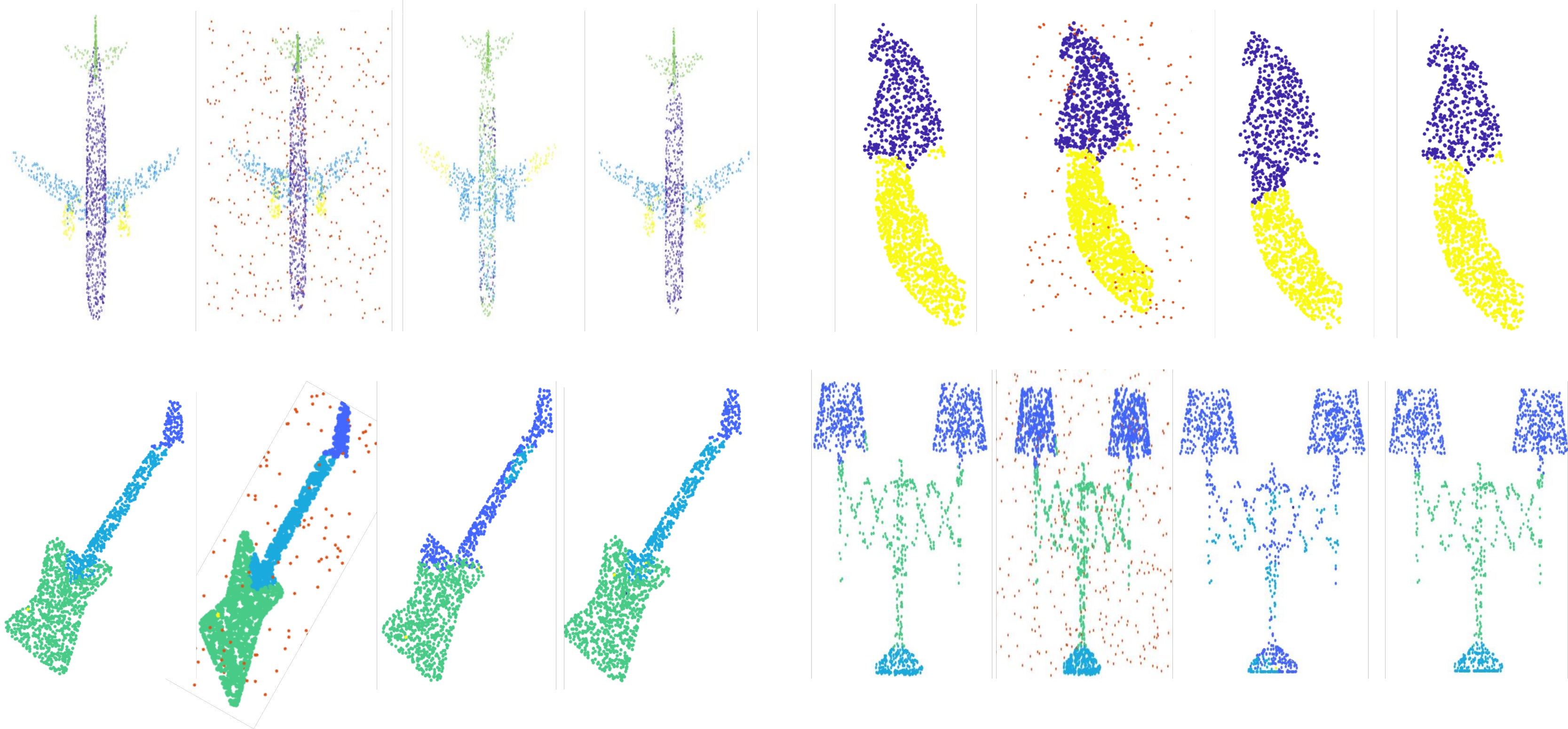}
\caption{ Four samples of ShapeNet part dataset showing the original objects, the same objects augmented by outliers, PoinNet with max pooling segmentation results, and finally PoinNet with histogram pooling segmentation results (ours).}
\label{fig_seg}
\end{figure*}

We augmented the testing part of the ShapeNet dataset with different ratios of outliers to test the performance of PointNet with different pooling operations.
Table~\ref{tab-seg} shows the results of mean IoU (mIoU) and per-category scores (note that mIoU is calculated only for inliers). When there are no outliers, histogram pooling achieves 77\% mIoU, max pooling scores 83\%,  while RANSAC and TQ score 81\% and 83\% respectively. However when outliers are added, max pooling mIoU drops, significantly. In contrast, histogram, RANSAC, and TQ mIoU almost remains constant for different outlier ratios. Figure~\ref{fig_seg} shows instances of segmented objects with outliers. As can be seen, segments with histogram pooling is almost similar to the original object segments, while with max pooling, many segments are miss-classified.

The above results indicate that the global features generated by histogram pooling carried robust information about the shape of the object. Unlike max pooling, histogram, RANSAC, and TQ pooling operations enabled the decoder part of the network to segment the objects correctly. 

\begin{table*}[h]
\fontsize{10}{12}\selectfont 
\caption{Segmentation results on ShapeNet part dataset. We compare PointNet vanilla with max, RANSAC, TQ and histogram pooling. The results show the importance of using robust pooling over max.}
\resizebox{\textwidth}{!}{%
\label{tab-seg}

\begin{tabular}{c|l|c|llllllllllllllll}
\hline
\begin{tabular}[c]{@{}c@{}}outl\\ \%\end{tabular} & \multicolumn{1}{c|}{Pool} & mean & \multicolumn{1}{c}{Airo} & \multicolumn{1}{c}{Bag} & \multicolumn{1}{c}{Cap} & \multicolumn{1}{c}{Car} & \multicolumn{1}{c}{Chair} & \multicolumn{1}{c}{\begin{tabular}[c]{@{}c@{}}Ear\\ phone\end{tabular}} & \multicolumn{1}{c}{Guitar} & \multicolumn{1}{c}{Knife} & \multicolumn{1}{c}{Lamp} & \multicolumn{1}{c}{Laptop} & \multicolumn{1}{c}{\begin{tabular}[c]{@{}c@{}}Motor\\ bike\end{tabular}} & \multicolumn{1}{c}{Mug} & \multicolumn{1}{c}{Pistol} & \multicolumn{1}{c}{Rocket} & \multicolumn{1}{c}{\begin{tabular}[c]{@{}c@{}}Skate\\ board\end{tabular}} & \multicolumn{1}{c}{Table} \\ \hline

\hline
\multicolumn{1}{c|}{\multirow{4}{*}{0}}  & max & \textbf{0.83} & 0.83 & 0.72 & 0.74 & 0.74 & 0.89 & 0.71 & 0.91 & 0.85 & 0.80 & 0.95 & 0.64 & 0.92 & 0.80 & 0.53 & 0.73 & 0.82 \\
\multicolumn{1}{c|}{}                    & HS  & 0.78 & 0.77 & 0.59 & 0.67 & 0.66 & 0.84 & 0.66 & 0.88 & 0.80 & 0.76 & 0.93 & 0.45 & 0.83 & 0.69 & 0.41 & 0.65 & 0.77 \\
\multicolumn{1}{c|}{}                    & RN  & 0.81 & 0.80 & 0.72 & 0.77 & 0.70 & 0.87 & 0.72 & 0.89 & 0.83 & 0.79 & 0.94 & 0.57 & 0.89 & 0.78 & 0.53 & 0.69 & 0.80 \\
\multicolumn{1}{c|}{}                    & TQ  & 0.83 & 0.82 & 0.74 & 0.76 & 0.72 & 0.88 & 0.72 & 0.90 & 0.84 & 0.80 & 0.95 & 0.60 & 0.92 & 0.81 & 0.54 & 0.70 & 0.81 \\ \hline
\multicolumn{1}{c|}{\multirow{4}{*}{5}}  & max & 0.42 & 0.25 & 0.46 & 0.49 & 0.35 & 0.61 & 0.37 & 0.49 & 0.77 & 0.50 & 0.49 & 0.26 & 0.62 & 0.27 & 0.24 & 0.26 & 0.28 \\
\multicolumn{1}{c|}{}                    & HS  & 0.78 & 0.76 & 0.58 & 0.66 & 0.66 & 0.85 & 0.66 & 0.88 & 0.79 & 0.76 & 0.93 & 0.44 & 0.83 & 0.69 & 0.43 & 0.65 & 0.77 \\
\multicolumn{1}{c|}{}                    & RN  & 0.81 & 0.80 & 0.72 & 0.77 & 0.70 & 0.87 & 0.72 & 0.89 & 0.83 & 0.79 & 0.94 & 0.57 & 0.89 & 0.78 & 0.53 & 0.69 & 0.80 \\
\multicolumn{1}{c|}{}                    & TQ  & \textbf{0.83} & 0.82 & 0.74 & 0.76 & 0.72 & 0.88 & 0.72 & 0.90 & 0.83 & 0.79 & 0.95 & 0.60 & 0.92 & 0.81 & 0.53 & 0.7  & 0.81 \\ \hline
\multicolumn{1}{c|}{\multirow{4}{*}{20}} & max & 0.34 & 0.21 & 0.45 & 0.53 & 0.25 & 0.47 & 0.27 & 0.42 & 0.77 & 0.48 & 0.39 & 0.24 & 0.54 & 0.17 & 0.23 & 0.28 & 0.18 \\
\multicolumn{1}{c|}{}                    & HS  & 0.77 & 0.77 & 0.62 & 0.71 & 0.65 & 0.85 & 0.66 & 0.88 & 0.79 & 0.76 & 0.93 & 0.44 & 0.81 & 0.69 & 0.45 & 0.66 & 0.77 \\
\multicolumn{1}{c|}{}                    & RN  & 0.81 & 0.80 & 0.70 & 0.75 & 0.68 & 0.87 & 0.72 & 0.87 & 0.82 & 0.78 & 0.95 & 0.56 & 0.89 & 0.78 & 0.52 & 0.67 & 0.79 \\
\multicolumn{1}{c|}{}                    & TQ  & \textbf{0.82} & 0.80 & 0.72 & 0.75 & 0.70 & 0.88 & 0.76 & 0.89 & 0.81 & 0.80 & 0.94 & 0.56 & 0.91 & 0.80 & 0.49 & 0.67 & 0.80 \\ \hline

\end{tabular}}%
\end{table*}

\section{Sensitivity analysis} \label{sec:ABLATION STUDY}

In this section, we evaluate the performance of the proposed pooling layer as part of the PointNet and DGCNN methods. The study also includes the sensitivity analysis of the histogram bin size on the classification accuracy. In these experiments, the data is augmented by 10\% additive noise and 50\% outliers. The performance of histogram pooling layer within different network structures is shown in table~\ref{Tab:abl}. The first row shows the classification accuracy of the PointNet when it uses the two transformation networks that are designed to estimate rotation and translation - referred to as PointNet(1). The last two rows show the classification accuracy of PointNet vanilla (PointNet without transformation networks) when histogram pooling layer is used - referred to PointNet(2). It is somewhat surprising to note that the classification accuracy of PointNet decreases by using transformation networks. The transformation networks appear to be overly sensitive to data augmentation. The last row shows that using point normals, in PointNet(2) achieves higher classification accuracy for clean data, while its robustness against outliers and noise is less than the case when it uses point coordinates.

The second row shows the classification accuracy of DGCNN when histogram pooling layer is used - referred to as DGCNN(1). The method showed low robustness to outliers and noise compared to PointNet(2). In the third row, we modified the convolution layers of the DGCNN(1) to include only neighbouring points within a certain radius (0.25 for the first convolution layer, and 2 for the rest of the convolution layers). This is called DGCNN(2) and its robustness to data augmentation is enhanced by around 10\%. In the fourth row, we modified DGCNN to include only two convolution layers, which is called DGCNN(3). The results show that using only two convolution layers achieves the highest robustness to outlier and noise augmentation. 

 Figure~\ref{Tab:abl} shows the classification accuracy of the histogram pooling under several inlier thresholds (the threshold shown is the figure is half of the bin size). As can be seen from the figure, setting the threshold between 0.13-0.15 provides the highest robustness to outliers, while the classification accuracy of clean and noisy data remains constant.

\begin{table}[]
 \centering
\caption{classification performance on ModelNet40 with \textbf{histogram} pooling, clean: clean objects, outl: objects with 50\% outliers, noise: objects with 10\% noise}
\label{Tab:abl}
 \fontsize{10}{12}\selectfont 
\begin{tabular}{|c|c|c|c|c|}

\hline
Method                                                                      & Input                          & clean   & outl  & noise \\ \hline \hline
PointNet(1)                                                                    & \multirow{3}{*}{p}  & 0.69 & 0.58 & 0.63  \\ \cline{1-1} \cline{3-5} 
DGCNN(1)                                                                                        &                       & 0.84 & 0.39 & 0.3  \\DGCNN(2)                                                                                       &                       & 0.85 & 0.51 & 0.39 
\\DGCNN(3)                                                                                       &                       &  0.85 & \textbf{0.70} & 0.63   \\

\cline{1-1} \cline{3-5} 
\multirow{2}{*}{\begin{tabular}[c]{@{}c@{}}PointNet(2) \end{tabular}}                     &                       & 0.84 & 0.69  & \textbf{ 0.70}  \\ \cline{2-2} \cline{3-5} 
                                                                            & P+n                                      &\textbf{ 0.86} & 0.60  & 0.13  \\ \hline
\end{tabular}
\end{table}

\begin{figure}[]
\centering
\includegraphics[scale=.33]{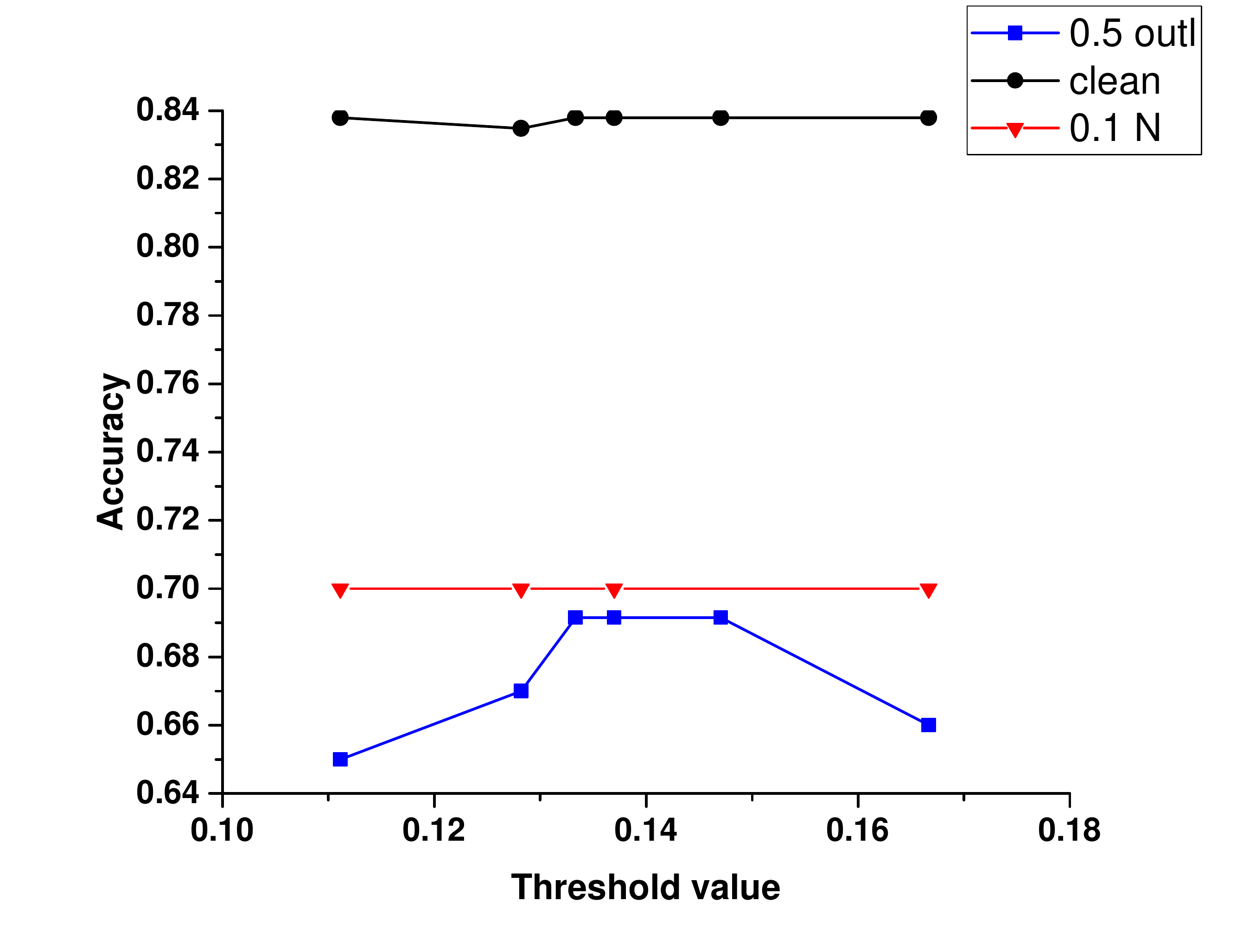}
\caption{Classification accuracy versus outliers.}
\label{fig_thresh}
\end{figure}

\hfill mds
 
\hfill April, 2021

\section{Limitation and future work}

The above results show that the robustness of point cloud classification networks can be significantly improved by using RANSAC or histogram pooling layers. PointNet with any of these pooling layers can tolerate large number of outliers and noise levels compared to the network with the max pooling layer. The only shortcoming of the robust pooling operations is the requirement of setting their thresholds. Moreover, RANSAC memory requirement grows rapidly with data size, which could limit its usage in applications with large point cloud datasets. Despite those shortcomings, both methods showed promising results and can open a window for future improvement in this area.





\section{Conclusion}
\label{sec:con}

We presented two pooling operations that are robust to data augmentation. The proposed pooling layers use histogram and RANSAC algorithm to look for clusters in data as clusters are indicatives of models. We tested those pooling layers with frameworks such as Point-based and graph based neural networks. Compared to max pooling, our results showed that the robustness of the proposed frameworks is significantly higher. When comparing our proposed pooling layers with robust state-of-the-art methods such as m-estimators, our histogram pooling was much faster and significantly more robust to outliers, with comparable robustness to noise and random point dropout. Compared to PL-Net3D, our histogram pooling was also significantly faster and more robust to noise and random point dropout, while we achieve similar robustness to outliers.

\appendices
\section{The confusion matrix of the  ScanObjectNN dataset}

\begin{figure}
    \centering
    \begin{tabular}{ccc}
        \includegraphics[width=0.15\textwidth]{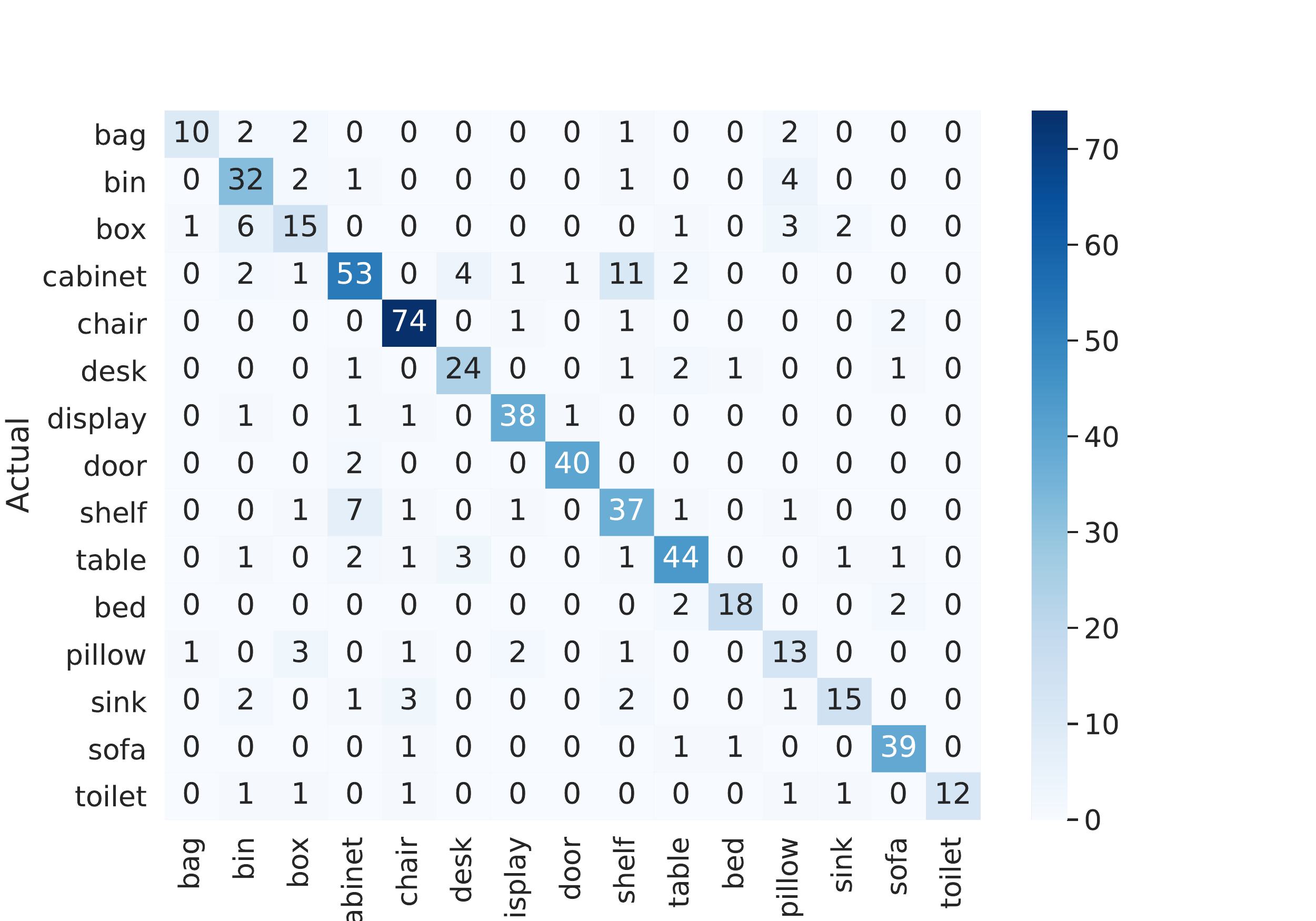} & \includegraphics[width=0.15\textwidth]{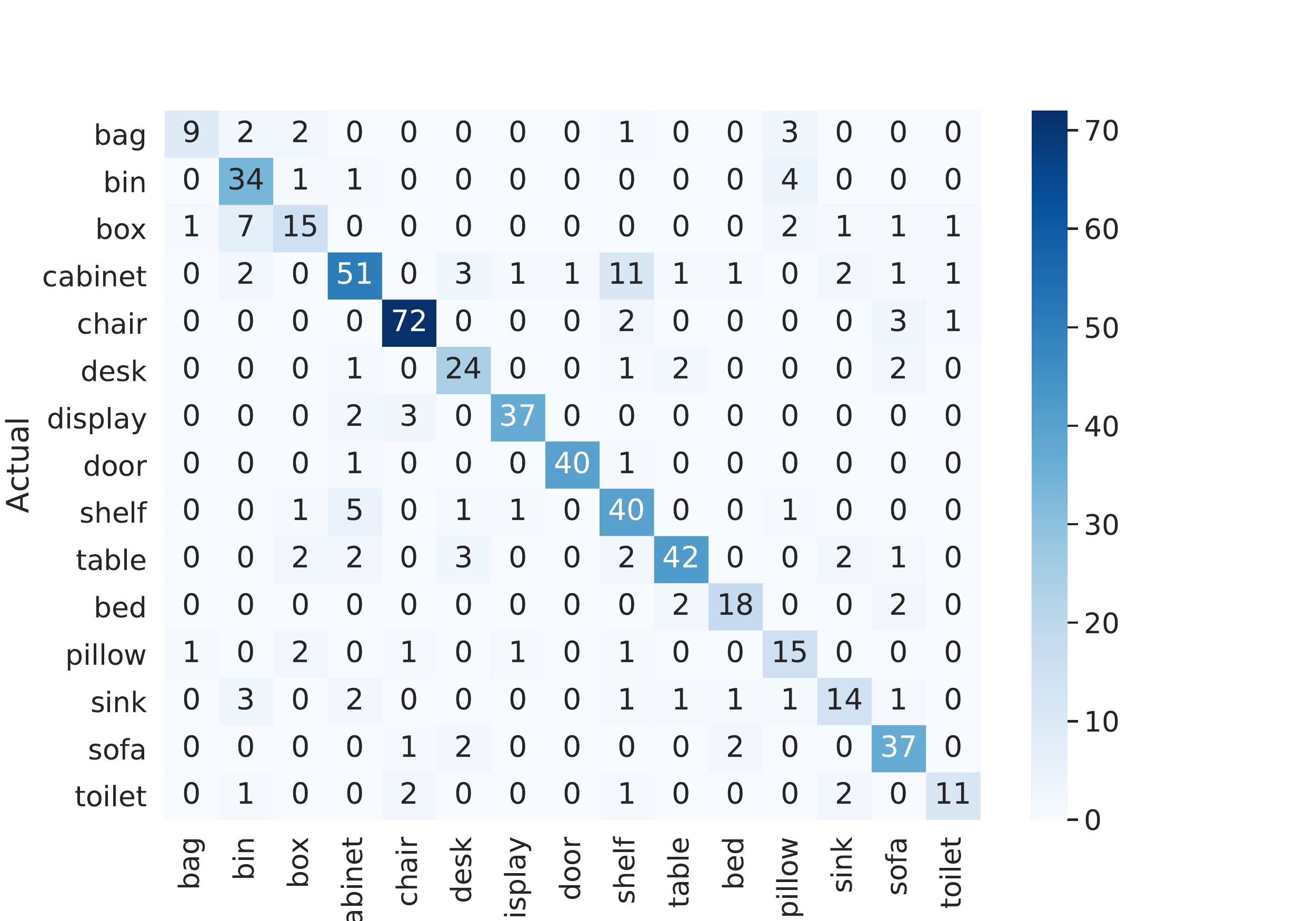}
        & \includegraphics[width=0.15\textwidth]{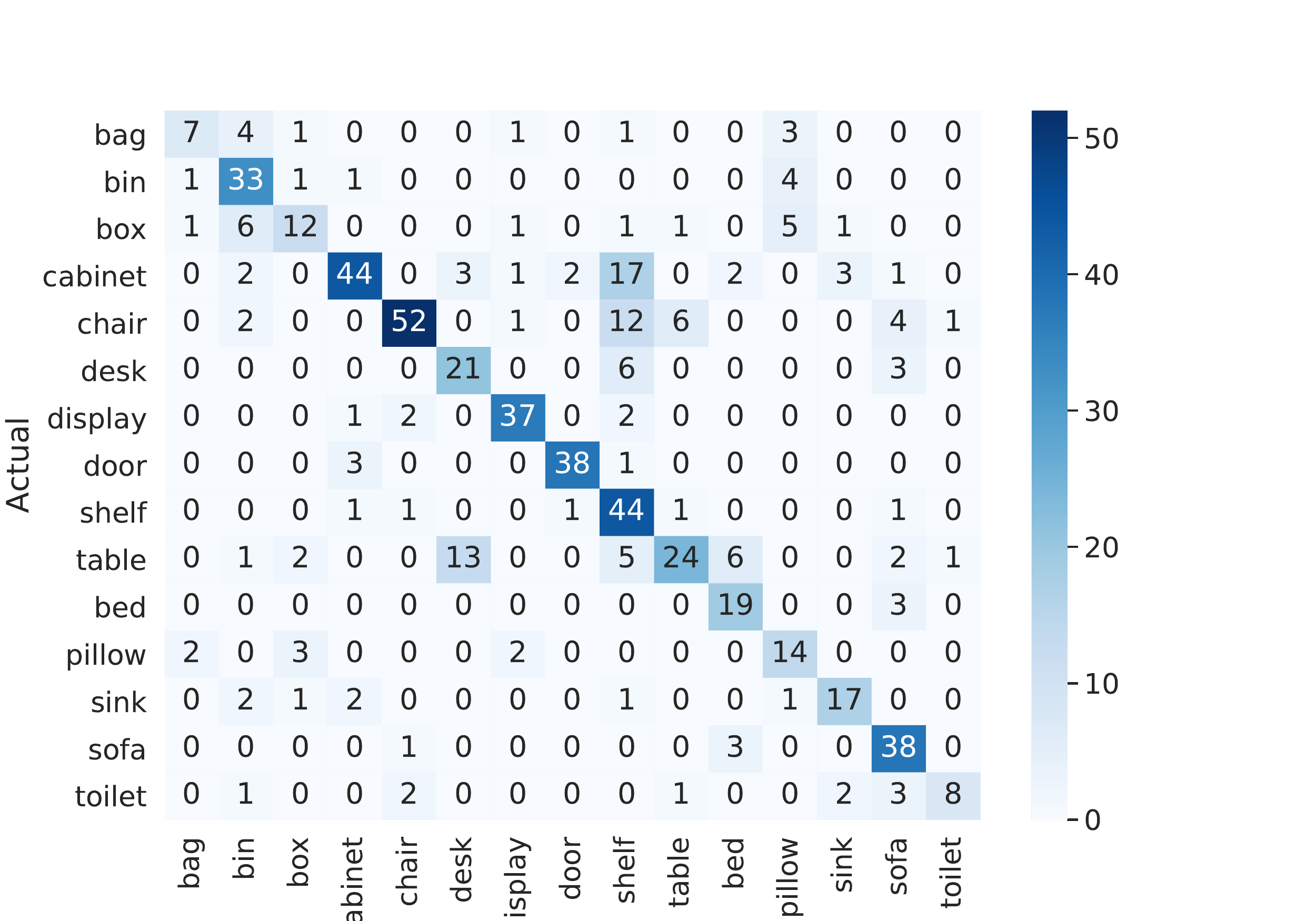}\\
        \small (a) & \small (b) & \small (c)
    \end{tabular}
    \captionsetup{singlelinecheck=off}
    \caption{The confusion matrix of the  ScanObjectNN dataset for PoinNet with histogram pooling for (a) clean object, (b) same object with 200 BG, and (c) same object with all BG points reported in the dataset.}
    \label{fig_conhs}
\end{figure}

\begin{figure}
    \centering
    \begin{tabular}{ccc}
        \includegraphics[width=0.15\textwidth]{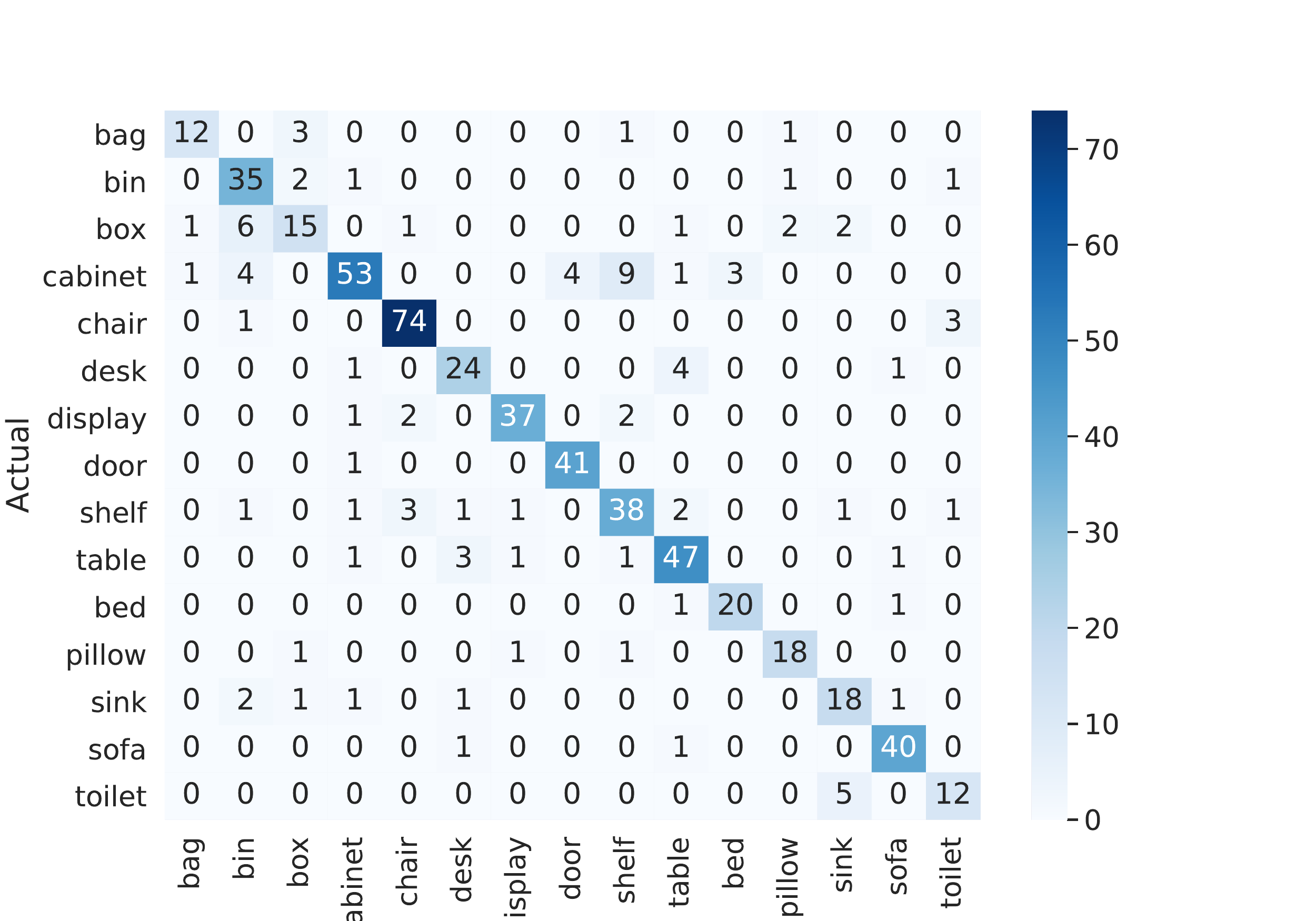} & \includegraphics[width=0.15\textwidth]{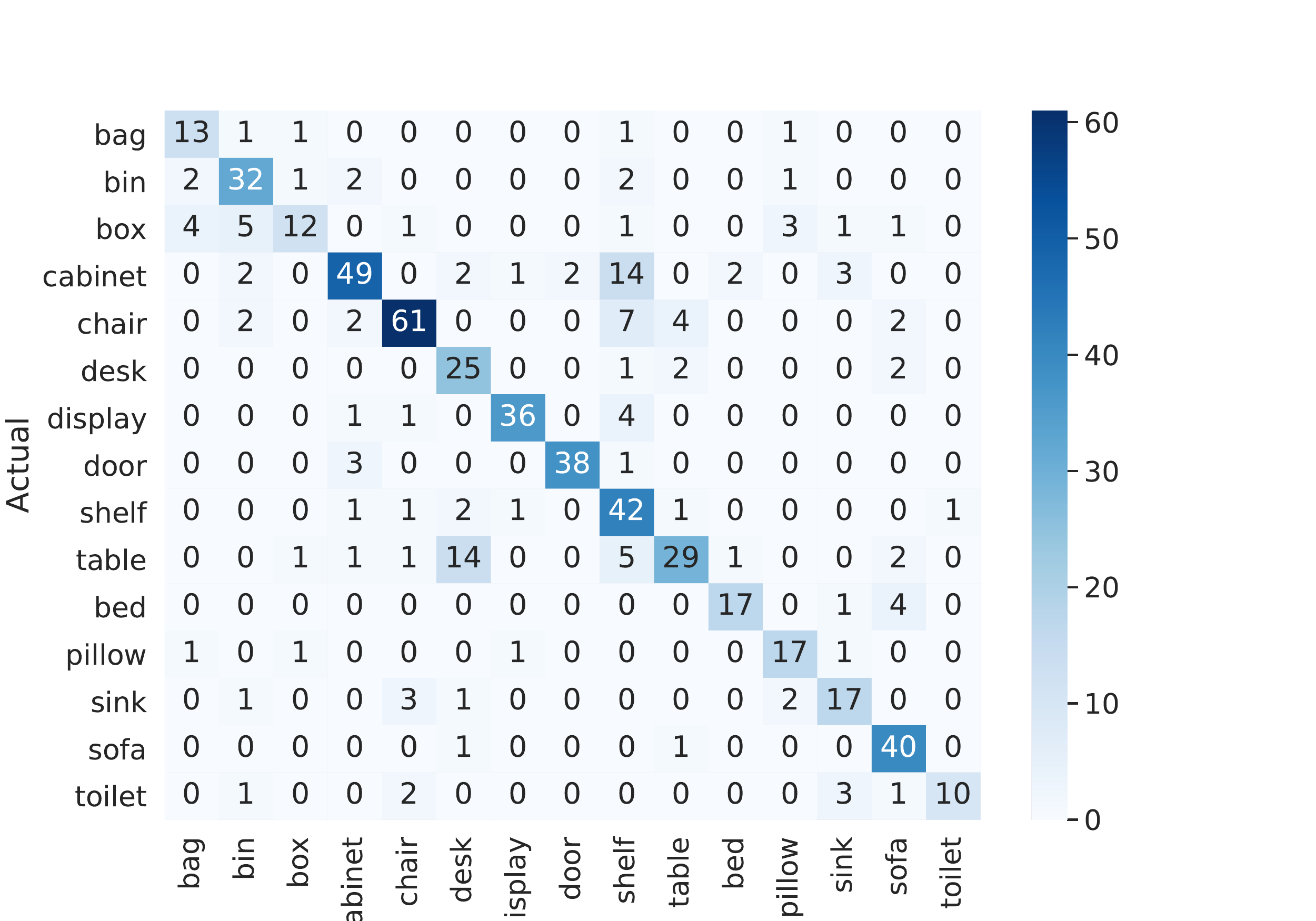}
        & \includegraphics[width=0.15\textwidth]{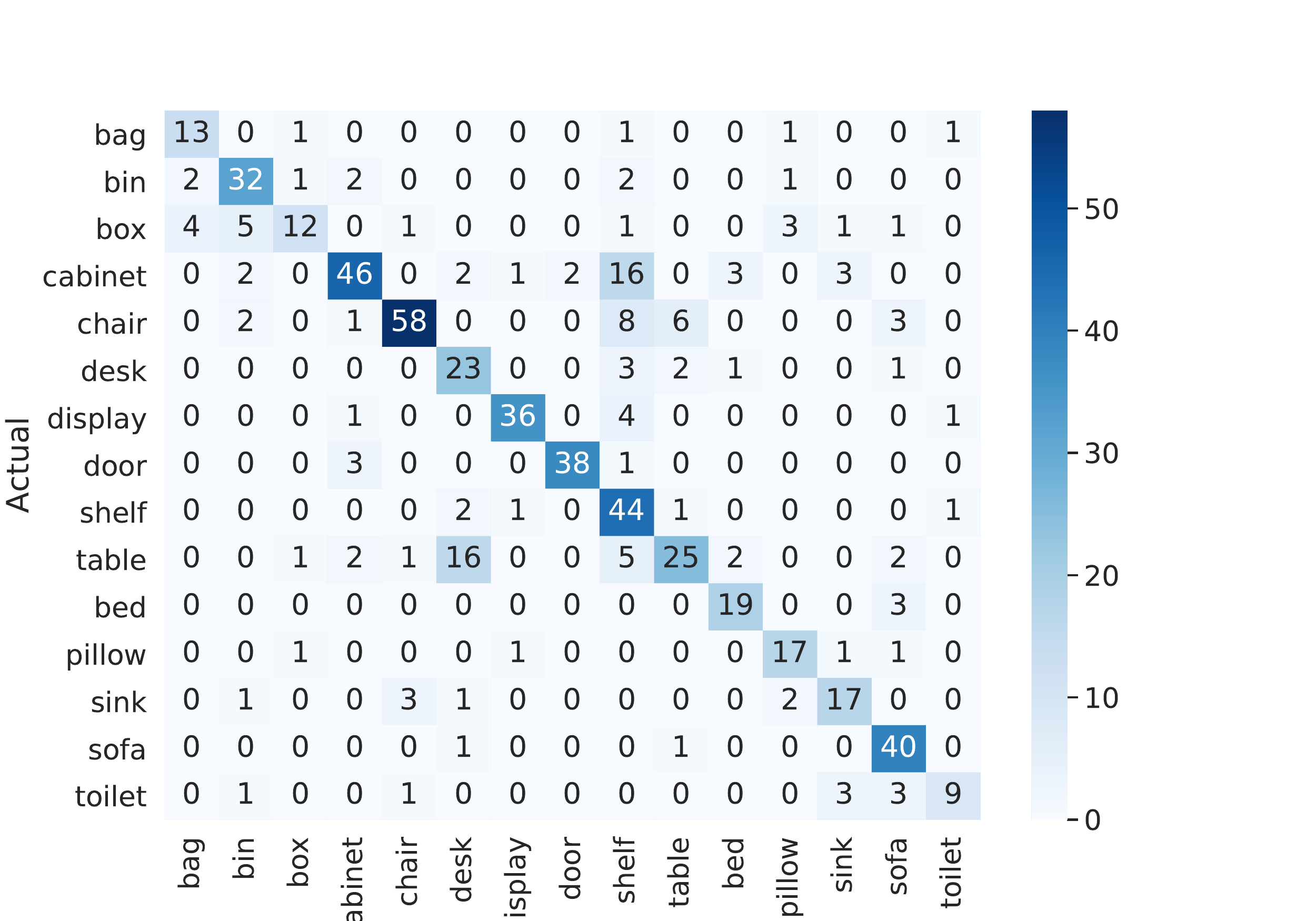}\\
        \small (a) & \small (b) & \small (c)
    \end{tabular}
    \captionsetup{singlelinecheck=off}
    \caption{The confusion matrix of the  ScanObjectNN dataset for PoinNet with max pooling for (a) clean object, (b) same object with 200 BG, and (c) same object with all BG points reported in the dataset.}
    \label{fig:conmmx}
\end{figure}

We show the confusion matrix of the  ScanObjectNN dataset for PoinNet with histogram and max pooling in figure~\ref{fig_conhs} and figure~\ref{fig:conmmx} respectively. Comparing both figures for objects with background data indicates that both methods misclassify tables to be desks due to the high similarity between both objects when background data exists.  

\ifCLASSOPTIONcaptionsoff
  \newpage
\fi



%

\bibliographystyle{IEEEtran} 

\bibliography{ref}

\begin{thebibliography}{10}
\providecommand{\url}[1]{#1}
\csname url@samestyle\endcsname
\providecommand{\newblock}{\relax}
\providecommand{\bibinfo}[2]{#2}
\providecommand{\BIBentrySTDinterwordspacing}{\spaceskip=0pt\relax}
\providecommand{\BIBentryALTinterwordstretchfactor}{4}
\providecommand{\BIBentryALTinterwordspacing}{\spaceskip=\fontdimen2\font plus
\BIBentryALTinterwordstretchfactor\fontdimen3\font minus
  \fontdimen4\font\relax}
\providecommand{\BIBforeignlanguage}[2]{{%
\expandafter\ifx\csname l@#1\endcsname\relax
\typeout{** WARNING: IEEEtran.bst: No hyphenation pattern has been}%
\typeout{** loaded for the language `#1'. Using the pattern for}%
\typeout{** the default language instead.}%
\else
\language=\csname l@#1\endcsname
\fi
#2}}
\providecommand{\BIBdecl}{\relax}
\BIBdecl

\bibitem{wu20153d}
Z.~Wu, S.~Song, A.~Khosla, F.~Yu, L.~Zhang, X.~Tang, and J.~Xiao, ``3d
  shapenets: A deep representation for volumetric shapes,'' in
  \emph{Proceedings of the IEEE conference on computer vision and pattern
  recognition}, 2015, pp. 1912--1920.

\bibitem{qi2017pointnet}
C.~R. Qi, H.~Su, K.~Mo, and L.~J. Guibas, ``Pointnet: Deep learning on point
  sets for 3d classification and segmentation,'' in \emph{Proceedings of the
  IEEE conference on computer vision and pattern recognition}, 2017, pp.
  652--660.

\bibitem{esteves2018learning}
C.~Esteves, C.~Allen-Blanchette, A.~Makadia, and K.~Daniilidis, ``Learning so
  (3) equivariant representations with spherical cnns,'' in \emph{Proceedings
  of the European Conference on Computer Vision (ECCV)}, 2018, pp. 52--68.

\bibitem{su2015multi}
H.~Su, S.~Maji, E.~Kalogerakis, and E.~Learned-Miller, ``Multi-view
  convolutional neural networks for 3d shape recognition,'' in
  \emph{Proceedings of the IEEE international conference on computer vision},
  2015, pp. 945--953.

\bibitem{klokov2017escape}
R.~Klokov and V.~Lempitsky, ``Escape from cells: Deep kd-networks for the
  recognition of 3d point cloud models,'' in \emph{Proceedings of the IEEE
  International Conference on Computer Vision}, 2017, pp. 863--872.

\bibitem{Ramasinghe}
S.~Ramasinghe, S.~Khan, N.~Barnes, and S.~Gould, ``Representation learning on
  unit ball with 3d roto-translational equivariance,'' \emph{International
  Journal of Computer Vision}, pp. 1--23, 2019.

\bibitem{qi2017pointnet++}
C.~R. Qi, L.~Yi, H.~Su, and L.~J. Guibas, ``Pointnet++: Deep hierarchical
  feature learning on point sets in a metric space,'' in \emph{Advances in
  neural information processing systems}, 2017, pp. 5099--5108.

\bibitem{wang2019dynamic}
Y.~Wang, Y.~Sun, Z.~Liu, S.~E. Sarma, M.~M. Bronstein, and J.~M. Solomon,
  ``Dynamic graph cnn for learning on point clouds,'' \emph{ACM Transactions on
  Graphics (TOG)}, vol.~38, no.~5, pp. 1--12, 2019.

\bibitem{siddiqi2008retrieving}
K.~Siddiqi, J.~Zhang, D.~Macrini, A.~Shokoufandeh, S.~Bouix, and S.~Dickinson,
  ``Retrieving articulated 3-d models using medial surfaces,'' \emph{Machine
  vision and applications}, vol.~19, no.~4, pp. 261--275, 2008.

\bibitem{chang2015shapenet}
A.~X. Chang, T.~Funkhouser, L.~Guibas, P.~Hanrahan, Q.~Huang, Z.~Li,
  S.~Savarese, M.~Savva, S.~Song, H.~Su \emph{et~al.}, ``Shapenet: An
  information-rich 3d model repository,'' \emph{arXiv preprint
  arXiv:1512.03012}, 2015.

\bibitem{mukhaimar2019comparative}
A.~Mukhaimar, R.~Tennakoon, C.~Y. Lai, R.~Hoseinnezhad, and A.~Bab-Hadiashar,
  ``Comparative analysis of 3d shape recognition in the presence of data
  inaccuracies,'' in \emph{2019 IEEE International Conference on Image
  Processing (ICIP)}.\hskip 1em plus 0.5em minus 0.4em\relax IEEE, 2019, pp.
  2471--2475.

\bibitem{gould2019deep}
S.~Gould, R.~Hartley, and D.~Campbell, ``Deep declarative networks: A new
  hope,'' \emph{arXiv preprint arXiv:1909.04866}, 2019.

\bibitem{mukhaimar2019pl}
A.~Mukhaimar, R.~Tennakoon, C.~Y. Lai, R.~Hoseinnezhad, and A.~Bab-Hadiashar,
  ``Pl-net3d: Robust 3d object class recognition using geometric models,''
  \emph{IEEE Access}, vol.~7, pp. 163\,757--163\,766, 2019.

\bibitem{fischler1981random}
M.~A. Fischler and R.~C. Bolles, ``Random sample consensus: a paradigm for
  model fitting with applications to image analysis and automated
  cartography,'' \emph{Communications of the ACM}, vol.~24, no.~6, pp.
  381--395, 1981.

\bibitem{shi2015deeppano}
B.~Shi, S.~Bai, Z.~Zhou, and X.~Bai, ``Deeppano: Deep panoramic representation
  for 3-d shape recognition,'' \emph{IEEE Signal Processing Letters}, vol.~22,
  no.~12, pp. 2339--2343, 2015.

\bibitem{sfikas2017exploiting}
K.~Sfikas, T.~Theoharis, and I.~Pratikakis, ``Exploiting the panorama
  representation for convolutional neural network classification and
  retrieval.'' in \emph{3DOR}, 2017.

\bibitem{kanezaki2018rotationnet}
A.~Kanezaki, Y.~Matsushita, and Y.~Nishida, ``Rotationnet: Joint object
  categorization and pose estimation using multiviews from unsupervised
  viewpoints,'' in \emph{Proceedings of the IEEE Conference on Computer Vision
  and Pattern Recognition}, 2018, pp. 5010--5019.

\bibitem{yu2018multi}
T.~Yu, J.~Meng, and J.~Yuan, ``Multi-view harmonized bilinear network for 3d
  object recognition,'' in \emph{Proceedings of the IEEE Conference on Computer
  Vision and Pattern Recognition}, 2018, pp. 186--194.

\bibitem{johns2016pairwise}
E.~Johns, S.~Leutenegger, and A.~J. Davison, ``Pairwise decomposition of image
  sequences for active multi-view recognition,'' in \emph{Proceedings of the
  IEEE Conference on Computer Vision and Pattern Recognition}, 2016, pp.
  3813--3822.

\bibitem{bai2016gift}
S.~Bai, X.~Bai, Z.~Zhou, Z.~Zhang, and L.~Jan~Latecki, ``Gift: A real-time and
  scalable 3d shape search engine,'' in \emph{Proceedings of the IEEE
  Conference on Computer Vision and Pattern Recognition}, 2016, pp. 5023--5032.

\bibitem{sfikas2018ensemble}
K.~Sfikas, I.~Pratikakis, and T.~Theoharis, ``Ensemble of panorama-based
  convolutional neural networks for 3d model classification and retrieval,''
  \emph{Computers \& Graphics}, vol.~71, pp. 208--218, 2018.

\bibitem{su2018deeper}
J.-C. Su, M.~Gadelha, R.~Wang, and S.~Maji, ``A deeper look at 3d shape
  classifiers,'' in \emph{Proceedings of the European Conference on Computer
  Vision (ECCV)}, 2018, pp. 0--0.

\bibitem{kim2020triplanar}
E.~Y. Kim, S.~Y. Shin, S.~Lee, K.~J. Lee, K.~H. Lee, and K.~M. Lee, ``Triplanar
  convolution with shared 2d kernels for 3d classification and shape
  retrieval,'' \emph{Computer Vision and Image Understanding}, vol. 193, p.
  102901, 2020.

\bibitem{Wu2015}
Z.~Wu, S.~Song, A.~Khosla, F.~Yu, L.~Zhang, X.~Tang, and J.~Xiao, ``3d
  shapenets: A deep representation for volumetric shapes,'' in
  \emph{Proceedings of the IEEE conference on computer vision and pattern
  recognition}, 2015, pp. 1912--1920.

\bibitem{riegler2017octnet}
G.~Riegler, A.~Osman~Ulusoy, and A.~Geiger, ``Octnet: Learning deep 3d
  representations at high resolutions,'' in \emph{Proceedings of the IEEE
  Conference on Computer Vision and Pattern Recognition}, 2017, pp. 3577--3586.

\bibitem{Vishakh2016a}
V.~Hegde and R.~Zadeh, ``Fusionnet: 3d object classification using multiple
  data representations,'' \emph{arXiv preprint arXiv:1607.05695}, 2016.

\bibitem{wang2017cnn}
P.-S. Wang, Y.~Liu, Y.-X. Guo, C.-Y. Sun, and X.~Tong, ``O-cnn: Octree-based
  convolutional neural networks for 3d shape analysis,'' \emph{ACM Transactions
  on Graphics (TOG)}, vol.~36, no.~4, p.~72, 2017.

\bibitem{maturana2015voxnet}
D.~Maturana and S.~Scherer, ``Voxnet: A 3d convolutional neural network for
  real-time object recognition,'' in \emph{2015 IEEE/RSJ International
  Conference on Intelligent Robots and Systems (IROS)}.\hskip 1em plus 0.5em
  minus 0.4em\relax IEEE, 2015, pp. 922--928.

\bibitem{zhou2018voxelnet}
Y.~Zhou and O.~Tuzel, ``Voxelnet: End-to-end learning for point cloud based 3d
  object detection,'' in \emph{Proceedings of the IEEE Conference on Computer
  Vision and Pattern Recognition}, 2018, pp. 4490--4499.

\bibitem{xiang2019novel}
B.~Xiang, J.~Tu, J.~Yao, and L.~Li, ``A novel octree-based 3-d fully
  convolutional neural network for point cloud classification in road
  environment,'' \emph{IEEE Transactions on Geoscience and Remote Sensing},
  2019.

\bibitem{Chen_2019_CVPR}
C.~Chen, G.~Li, R.~Xu, T.~Chen, M.~Wang, and L.~Lin, ``Clusternet: Deep
  hierarchical cluster network with rigorously rotation-invariant
  representation for point cloud analysis,'' in \emph{The IEEE Conference on
  Computer Vision and Pattern Recognition (CVPR)}, June 2019.

\bibitem{Zhang2019RotationIC}
Z.~Zhang, B.-S. Hua, D.~W. Rosen, and S.-K. Yeung, ``Rotation invariant
  convolutions for 3d point clouds deep learning,'' \emph{ArXiv}, vol.
  abs/1908.06297, 2019.

\bibitem{li2018so}
J.~Li, B.~M. Chen, and G.~Hee~Lee, ``So-net: Self-organizing network for point
  cloud analysis,'' in \emph{Proceedings of the IEEE conference on computer
  vision and pattern recognition}, 2018, pp. 9397--9406.

\bibitem{liu2019densepoint}
Y.~Liu, B.~Fan, G.~Meng, J.~Lu, S.~Xiang, and C.~Pan, ``Densepoint: Learning
  densely contextual representation for efficient point cloud processing,'' in
  \emph{Proceedings of the IEEE International Conference on Computer Vision},
  2019, pp. 5239--5248.

\bibitem{liu2019relation}
Y.~Liu, B.~Fan, S.~Xiang, and C.~Pan, ``Relation-shape convolutional neural
  network for point cloud analysis,'' in \emph{Proceedings of the IEEE
  Conference on Computer Vision and Pattern Recognition}, 2019, pp. 8895--8904.

\bibitem{liu2019point}
Z.~Liu, H.~Tang, Y.~Lin, and S.~Han, ``Point-voxel cnn for efficient 3d deep
  learning,'' in \emph{Advances in Neural Information Processing Systems},
  2019, pp. 965--975.

\bibitem{leroy1987robust}
A.~M. Leroy and P.~J. Rousseeuw, ``Robust regression and outlier detection,''
  \emph{rrod}, 1987.

\bibitem{uy2019revisiting}
M.~A. Uy, Q.-H. Pham, B.-S. Hua, T.~Nguyen, and S.-K. Yeung, ``Revisiting point
  cloud classification: A new benchmark dataset and classification model on
  real-world data,'' in \emph{Proceedings of the IEEE/CVF International
  Conference on Computer Vision}, 2019, pp. 1588--1597.

\bibitem{yi2016scalable}
L.~Yi, V.~G. Kim, D.~Ceylan, I.-C. Shen, M.~Yan, H.~Su, C.~Lu, Q.~Huang,
  A.~Sheffer, and L.~Guibas, ``A scalable active framework for region
  annotation in 3d shape collections,'' \emph{ACM Transactions on Graphics
  (ToG)}, vol.~35, no.~6, pp. 1--12, 2016.

\end{thebibliography}

%




\end{document}